\documentclass{AISE}  
\usepackage{multirow}
\usepackage{adjustbox}
\hypersetup{breaklinks=true}
\usepackage{amsmath,amssymb,amsfonts}
\usepackage{graphicx}
\usepackage{subcaption}
\usepackage{textcomp}
\usepackage{xcolor}
\usepackage{float}

\renewcommand{\Issue}{31}
\renewcommand{\Ciation}{*** ***} 
\renewcommand{\Editor}{*** ***}

\title{Resilient Class-Incremental Learning: on the Interplay of Drifting, Unlabelled and Imbalanced Data Streams} 

\author{Jin Li$^{1,2}$, Kleanthis Malialis$^{1}$, Marios Polycarpou$^{1,2,*}$ \\[0.5em]
	\textit{\small $^{1}$ KIOS Research and Innovation Center of Excellence, University of Cyprus, Cyprus\\[0.5em]
            \small $^{2}$ Department of Electrical and Computer Engineering, University of Cyprus, Cyprus 
               } 
}

\renewcommand{\Abstract}{In today's connected world, the generation of massive streaming data across diverse domains has become commonplace. In the presence of concept drift, class imbalance, label scarcity, and new class emergence, they jointly degrade representation stability, bias learning toward outdated distributions, and reduce the resilience and reliability of detection in dynamic environments. This paper proposes SCIL (Streaming Class-Incremental Learning) to address these challenges. The SCIL framework integrates an autoencoder (AE) with a multi-layer perceptron for multi-class prediction, uses a dual-loss strategy (classification and reconstruction) for prediction and new class detection, employs corrected pseudo-labels for online training, manages classes with queues, and applies oversampling to handle imbalance. The rationale behind the method's structure is elucidated through ablation studies and a comprehensive experimental evaluation is performed using both real-world and synthetic datasets that feature class imbalance, incremental classes, and concept drifts. Our results demonstrate that SCIL outperforms strong baselines and state-of-the-art methods. Based on our commitment to Open Science, we make our code and datasets available to the community.
}

\renewcommand{\Keywords}{Concept drift, Data stream mining, Class incremental learning, Class imbalance }

\renewcommand{\InforFoot}{\textbf{*Corresponding Author}:  \href{mailto:email@email.com}{mpolycar@ucy.ac.cy}. \\
\textbf{Financial Support}: This paper was supported in part by funding from the European Research Council (ERC) under grant agreement No. 951424 (Water Futures) and the European Union's Horizon 2020 research and Innovation programme under grant agreement No. 739551 (KIOS CoE). \\
\textbf{Conflict of Interest Disclosure}: The authors report there are no competing interests to declare. This is an open access article under the CC BY-NC-ND license (\url{http://creativecommons.org/licenses/by-nc-nd/4.0/}).}

\begin{document}
	\coverpage
    \thispagestyle{empty}

    

\section{Introduction}

\subsection{Motivation and Open challenges}
\lettrine[lines=3, findent=5pt, nindent=0pt]{I}{n} recent years, the proliferation of streaming data across domains such as critical infrastructures, industrial systems, healthcare, and Internet of Things (IoT) networks has created unprecedented opportunities for real-time monitoring, decision making, and intelligent automation. However, this shift toward continuously evolving data streams also poses several critical challenges.

\textbf{Non-stationary environments}: While streaming data is often assumed to be generated by a stationary process, real-world scenarios are typically non-stationary due to \textbf{concept drift}\upcite{gama2014survey}, which refers to temporal changes in data characteristics. It may arise from seasonal effects, incipient faults, or shifts in user behavior. Drift can appear in different forms: abrupt drifts occur suddenly, incremental drifts evolve gradually through intermediate concepts, and recurrent drifts reintroduce previously observed patterns.

\textbf{Label unavailability}: 
Classification systems rely on labeled data but face practical limitations: labeling is often costly or infeasible (e.g., for minority classes like sensor faults), and streaming environments prevent timely annotation. The problem intensifies with the emergence of unseen classes, which lack labels entirely.

\textbf{Class imbalance}: Infrequent or anomalous events can degrade model performance by biasing predictions toward the normal class\upcite{he2008learning}. The challenge is amplified in streaming settings\upcite{wang2018systematic}, where normal operation forms the majority class while multiple anomaly types constitute minority classes.

\textbf{Incremental new classes}: 
In streaming environments, the emergence of unseen classes creates the incremental new class problem. Accurate detection is crucial in safety-critical domains (e.g., predictive maintenance, network security\upcite{Lichman2013}), where emerging faults might otherwise go undetected. Incorporating new classes introduces further challenges: updating models solely with new-class data can overwrite prior knowledge, causing catastrophic forgetting\upcite{zhou2024class}—a core challenge in class-incremental learning.

\subsection{Interplay of challenges and illustrative applications}

These challenges interact in complex ways: concept drift blurs class boundaries, causing misclassification; catastrophic forgetting overwrites prior knowledge when adapting to new classes; class imbalance biases decisions toward outdated boundaries; label scarcity prevents error correction; and inadequate new-class detection leads to overconfident mispredictions of novel events. Therefore, it is important to obtain solutions in the presence of multiple challenges.

Across water systems\upcite{li2025online}, cybersecurity\upcite{sommer2010outside}, and industrial inspection\upcite{ge2025class}, common challenges emerge: significant class imbalance between rare anomalies (e.g., water contaminations, attacks, faults) and normal operations; concept drift from sensor degradation, network changes, or environmental variations; and novel classes from emerging threats or defects. While continuous adaptation is essential, it risks catastrophic forgetting of prior patterns, making reliable and resilient detection crucial to avoid severe consequences.

\subsection{Contributions}
To tackle these challenges, ideally a model must learn from unlabeled data, handle class imbalance, adapt to concept drift, and detect new classes. While previous studies address some of these challenges separately, in real-world applications, they do not occur separately or independently of one another. The objective of this paper is to develop a method that addresses all of these challenges, so that machine learning models are resilient and continue to perform reasoning in such applications. The main contributions of this study are as follows:
\begin{enumerate}

    \item We propose SCIL (Streaming Class-Incremental Learning), a resilient framework that combines an autoencoder (AE) and a multi-layer perceptron (MLP) for multi-class classification and  detection. The AE's latent representations feed into the MLP, forming a unified model trained with a dual-loss objective.

\item We propose a pseudo-label oversampling strategy for online training, integrated with a reliability-aware correction mechanism based on class density, scale, and distance. This approach enhances the quality of emerging-class samples for model updates while, through a dynamic replay-based storage mechanism, effectively adapting to non-stationary environments and mitigating catastrophic forgetting.

    \item We extensively evaluate SCIL on real-world and synthetic datasets through ablation studies and comparisons with state-of-the-art methods, demonstrating significant performance gains in class-imbalanced incremental new-class scenarios under nonstationary environments.
\end{enumerate}

The structure of the paper is as follows:  Sec.~\ref{sec:background} introduces the background and Sec.~\ref{sec:related} reviews related work. The proposed method is detailed in Sec.~\ref{sec:method}, followed by the experimental setup in Sec.~\ref{sec:exp_setup}. Sec.~\ref{sec:exp_results} provides an empirical analysis of the proposed approach along with comparative studies of various learning methods. Finally, concluding remarks are discussed in Sec.~\ref{sec:conclusion}. For the reproducibility of our results, the datasets used and our code are made publicly available to the community\footnote{https://github.com/Jin000001/SCIL}

\section{Background}\label{sec:background}
Online learning refers to a continuous process of data generation, where at each time step $t$, a set of examples $S^t = \{(\boldsymbol{x}^t_i, y^t_i)\}^M_{i=1}$ is provided, where $M$ is the number of examples in that batch. The data is usually drawn from a long or potentially unending sequence. When $M=1$, it is referred to as \textbf{one-by-one online learning}, which is the focus of this work. The examples are assumed to be drawn from a time-varying, unknown probability distribution $p^t(x, y)$, with $\boldsymbol{x}^t \in \mathbb{R}^d$ as a $d$-dimensional vector from the input space 
$X \subset \mathbb{R}^d$, and $y^t \in Y = \{0, 1, \dots, n\}$ as the class label. The \textbf{concept drift} between timesteps $i$ and $j$ can be formally defined as $p^{t_i}(x,y) \neq p^{t_j}(x,y)$ for $i \ne j$. In this study, we address the multi-class problem, where $n$ is the number of known minority classes (i.e., `majority' ($y=0$), `minority\_1' ($y=1$), `minority\_2' ($y=2$), \dots, `minority\_n' ($y=n$)). In this study, we define the \textbf{incremental new classes} as follows. During the offline learning phase at $t=0$, we have an initial set of classes: 
$Y^0 = \{0, 1, \dots, j\}$. At a random subsequent timestep $t_1$, $Y^{t_1} = \{0, 1, \dots, j+1\}$. Then, at another random timestep $t_2$, the class set is expanded to $Y^{t_2} = \{0, 1, \dots, j+2\}$. This process continues at random timesteps $t_3, t_4, \dots$, with $0 < t_1 < t_2 < t_3 < \cdots$. The \textbf{class imbalance} is defined as $\forall\, i \in \{1,2,\ldots,n\},\, p^{t}(y=i) \ll p^{t}(y=0)$, where $i$ denotes the minority classes. 
For example, in the case of the fault diagnosis problem, the `majority' represents the healthy case, while the minorities may represent different types of faults. In this study, we focus on the one-by-one online \textbf{unsupervised} learning paradigm, 
which operates without requiring any class labels, i.e., $S^t = (\boldsymbol{x}^t)$. The \textbf{incremental learning}\upcite{losing2018incremental} adopted in this study refers to the continuous adaptation of a model without full re-training. 
Instead, the model is updated iteratively, represented as $h^t = h^{t-1}.\mathrm{train}(\cdot)$.

\section{Related Work}\label{sec:related}

\subsection{Concept drift adaptation}\label{sec:conceptdrift}
Concept drift handling is categorized as passive or active\upcite{ditzler2015learning}. Passive methods use incremental learning for implicit adaptation, where models update iteratively ($h^t = h^{t-1}.train(\cdot)$)\upcite{losing2018incremental}. These include memory-based methods (sliding windows\upcite{liang2006fast,lazarescu2004using}) and ensemble methods (dynamic model collections like  Dealing with Drifts (DDD)\upcite{minku2011ddd} and Streaming Ensemble Algorithm (SEA)\upcite{street2001streaming}).

Active methods rely on explicit drift detection to trigger an adaptation mechanism. Two main categories of detection mechanisms have been explored: statistical tests (e.g., \cite{jaworski2020concept, friedrich2023unsupervised}) and threshold-based mechanisms (e.g., \cite{menon2020concept, greco2025unsupervised, zhao2023unsupervised}). Statistical tests analyze the statistical properties of the generated data, whereas threshold-based mechanisms track prediction errors and assess them against a set threshold.

\subsection{Online imbalanced learning}\label{sec:anodetec}

While class imbalance is well-studied in binary anomaly detection, multi-class settings have received less attention. Under certain conditions, binary methods can extend to multi-class, so we review representative online unsupervised binary approaches here; supervised methods are surveyed elsewhere\upcite{wang2018systematic, aguiar2024survey}. Online class-incremental learning follows in the next subsection.

Anomaly detection typically involves learning normal behavior and flagging deviations. Methods include: Local Outlier Factor (LOF)\upcite{breunig2000lof}, which detects outliers via local density; One-Class SVM (OC-SVM)\upcite{scholkopf2001estimating}, which learns a normal boundary; Gaussian Mixture Model (GMM)\upcite{veracini2009fully}, which models normality via Gaussian mixtures (though this assumption may not always hold); and Isolation Forest (iForest)\upcite{liu2008isolation}, which isolates anomalies via shorter tree paths. \cite{malialis2020online} adaptively maintains class balance.

Several deep learning methods address imbalanced binary classification in nonstationary environments: ARCUS\upcite{yoon2022adaptive} uses concept-driven inference and drift-aware updates for AE-based anomaly detection; MemStream\upcite{bhatia2022memstream} employs a denoising autoencoder with memory for drift handling; Streaming Autoencoder (SA)\upcite{dong2018threaded} uses an ensemble with incremental learning to distinguish drift from anomalies; and \cite{li2023autoencoder} proposes an autoencoder-based incremental method with drift detection.

\subsection{Online class incremental learning}\label{sec:mulcla}

In this section, we examine advanced weakly supervised techniques for multi-class classification under incremental class introduction. We group the methods into cluster-based and non-cluster-based approaches.

Clustering-based methods for evolving streams include: CPOCEDS\upcite{jafseer2024cpoceds}, which uses online K-Means with three phases and stores outliers as potential new class indicators; OLINDDA\upcite{spinosa2009novelty}, a pioneering cluster-based novelty detector using distance thresholds; MINAS\upcite{de2016minas}, which builds class representations via clusters and classifies unexplained instances as unknown (our method uses reconstruction loss instead); ensemble methods like KNNENS\upcite{zhang2022knnens} and OCGCD\upcite{park2024online} combining representation learning with clustering; UDOR\upcite{alfarisy2025towards} integrating self-supervised learning with cluster-based discovery; and ensemble approaches using homogeneous/heterogeneous clustering partitions\upcite{garcia2019ensemble}. However, class imbalance impacts clustering effectiveness\upcite{xuan2013exploring}. Other methods like StreamWNN\upcite{melgar2023novel} (time-series specific) and distributed anomaly detection\upcite{jain2021distributed} (requiring online labels) have limitations. Importantly, class imbalance and misclassification challenge cluster formation.

Some non-cluster-based approaches are proposed. Related methods include: LC-INC\upcite{zhou2021learning}, which uses prototype networks for novelty detection with few samples but lacks concept drift handling; SNDProb\upcite{carreno2022sndprob}, which models classes via Gaussian mixtures and identifies new classes via EM; SENCForest\upcite{mu2017classification}, which detects emerging classes via anomaly detection; \cite{gama2024fault}, a neural-symbolic approach for fault explanation requiring prior class knowledge; and \cite{jiang2024few}, which regularizes feature learning assuming few new classes per batch—often unrealistic online.

\subsection{Research Gaps}  
Most existing methods for addressing concept drift rely on supervised learning. In streaming data, concept drift may naturally occur and poses an additional challenge to new class detection methods\upcite{agrahari2024review}. Moreover, in multi-class and incremental-class settings, effectively managing both concept drift and class imbalance remains an open research problem\upcite{wang2018systematic}. While some works attempt to address both issues, they are limited to binary classification. Among the discussed methods, only LC-INC explicitly considers class imbalance, but it does not handle concept drift in an unsupervised manner. A few other methods focus on concept drift\upcite{garcia2019ensemble, jafseer2024cpoceds, de2016minas, carreno2022sndprob} without addressing class imbalance. 

Moreover, in the latest class-incremental learning (CIL) survey\upcite{zhou2024class}, the authors note that handling imbalance, weak supervision, and concept drift remains underexplored and is highlighted as a key direction for future work. Most current CIL methods are fully supervised and trained in offline settings.

In contrast to existing methods that typically address these challenges separately, this work proposes a unified framework that simultaneously considers concept drift, class imbalance, and the appearance of new classes. By adopting an unsupervised and streaming learning paradigm, the proposed method overcomes several key limitations of existing approaches, particularly their reliance on supervision and offline training in multi-class incremental settings.

\begin{figure*}[!h]
	\centering \includegraphics[scale=0.5]{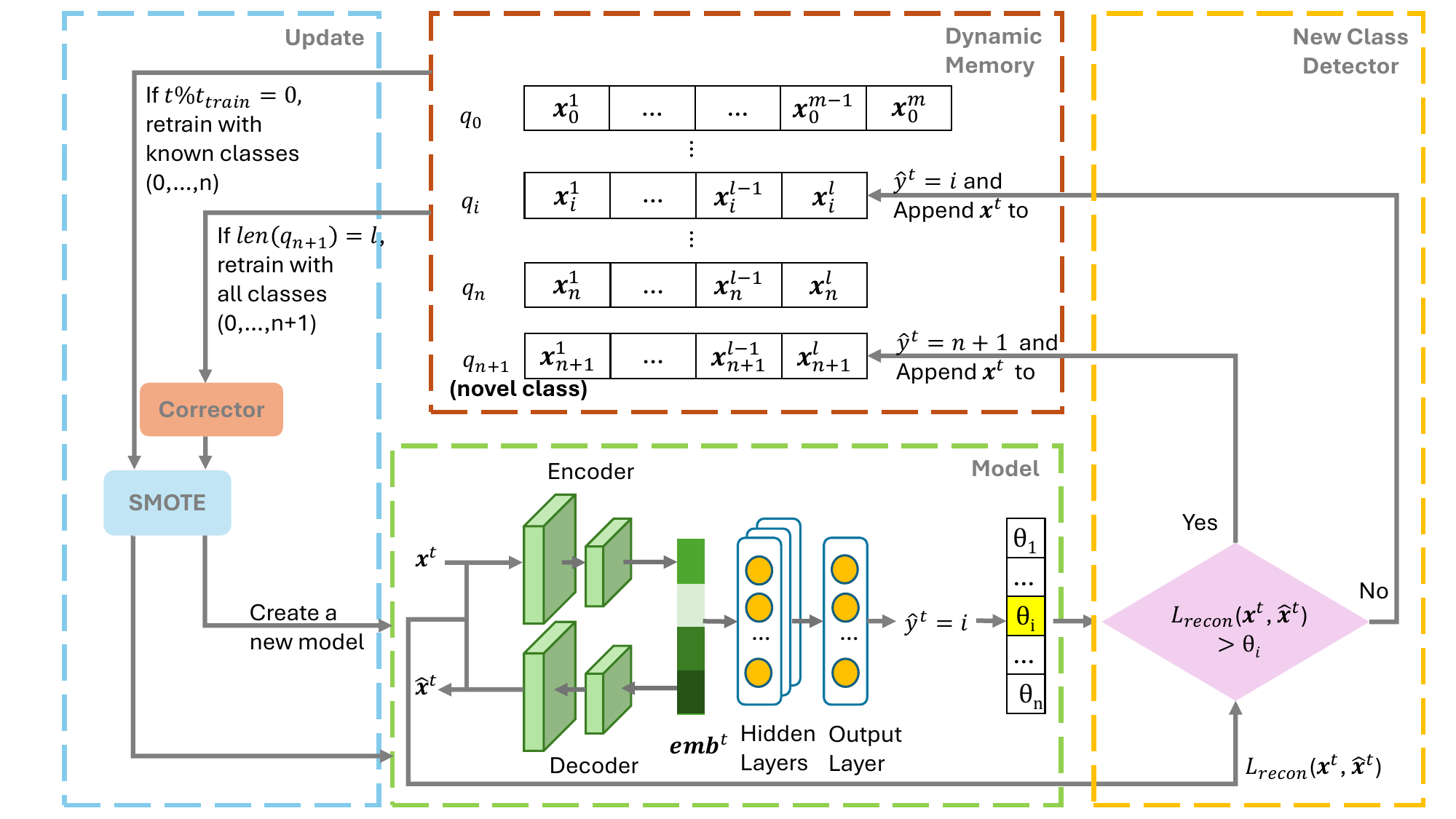}
	\caption{An overview of SCIL in block diagram representation.}
	\label{fig:SCIL}
\end{figure*}

\section{The SCIL Method}\label{sec:method}
In this study, we address an unsupervised, one-by-one learning problem. At each timestep $t$, an input example $\boldsymbol{x}^t$ arrives, and the model outputs a predicted label $\hat{y}^t$.

The overview of the proposed SCIL is illustrated in Fig.~\ref{fig:SCIL}. Specifically, we observe four components—`Dynamic Memory`, `Model`, `Update`, and `New Class Detector`—each outlined with dashed lines in distinct colors. The \textbf{Model} strategically integrates an AE with an MLP for multi-class classification and new class detection. The \textbf{New Class Detector} utilizes the model's prediction along with the AE’s reconstruction loss to identify novel classes. For each class, historical data is stored as queues in the \textbf{Dynamic Memory}, and novelty thresholds are established during offline learning. An instance is classified as either a sample of new class, which is stored separately, or a sample of seen classes, which is added to the corresponding class queue, based on a comparison between its reconstruction loss and the threshold of the predicted class. A new model is trained with resampled instances once the number of samples of the new class exceeds a predefined threshold. At specified intervals, the model undergoes incremental training, and its thresholds are updated to address concept drift. The \textbf{Update} process is illustrated in Fig.~\ref{fig:SCIL}. Since labels are unavailable during online training, pseudo-labels—previous model predictions—are used for updates. The corresponding pseudocode is provided in Algorithm 1. The detailed description of each part will be discussed in the following.

\begin{algorithm}[!h]
\small
\caption{SCIL}
\label{alg:method}
\KwIn{$t_{train}$: incremental training interval; $D_i$: labeled data for pre-training, $i \in \{0, 1, ... n\}$, where 0 represents majority class and $n$ is the number of known minority class; $m$: queue size of majority class; $l$: queue size of minority and new classes}
\textbf{Init:} \tcp*{time $t=0$}
$h = \text{train}(D_0, D_1, ..., D_n)$ \tcp*[l]{pretraining}
$\theta_0, \theta_1, ..., \theta_n$ according to Eq.~(\ref{eq:threshold_0}) and Eq.~(\ref{eq:threshold_i})\;
$q_0$ (capacity=$m$, init=$\{D_0\}$)\;
$q_{n+1}$ (capacity=$l$, init=None)\;
$q_1, ..., q_n$ (capacity=$l$, init=$\{D_1, ..., D_n\}$)\;

\BlankLine
\textbf{Start:}\;
\For{$t = 1, 2, ... \infty$}{
    receive instance $\boldsymbol{x}^t \in \mathbb{R}^d$\;
    predict $\hat{y}^t = h.\text{predict}(\boldsymbol{x}^t) = i$ \tcp*[l]{predicted as $i$}
    
    \eIf{$\text{loss}_{\text{recon}}(\boldsymbol{x}^t, \hat{\boldsymbol{x}}^t) > \theta_i$}{
        $\hat{y}^t = n+1$ \tcp*[l]{new class detection}
        $q_{n+1}.\text{append}(\boldsymbol{x}^t)$\;
    }{
        $q_i.\text{append}(\boldsymbol{x}^t)$\;
    }
    
    \If{$\text{len}(q_{n+1}) == l$}{
        $n = n+1$ \tcp*[l]{incremental class}
        $q_n = q_{n+1}$\;
        $q_1, ..., q_n = \text{Corr}(q_1, ..., q_n)$ \tcp*[l]{see correction mechanism in Sec.~\ref{sec:update}}
        $h = \text{train}(\text{SMOTE}(q_0, ... q_n))$ \tcp*[l]{new model}
    }
    
    \If{$t \% t_{\text{train}}$}{
        $h.\text{train}(\text{SMOTE}(q_0, ... q_n))$ \tcp*[l]{incremental learning}
        $\theta_0, \theta_1, ..., \theta_n$ updated according to Eq.~(\ref{eq:threshold_0}) and Eq.~(\ref{eq:threshold_i})\;
    }
}
\end{algorithm}

\subsection{Dynamic Memory}
\textbf{Initial labeled data}. In the proposed method, we assume \( n+1 \) classes in the historical data: one majority class \( 0\) and \( n \) minority classes \( (1, \dots, n) \). The labeled initial data \( D_0 \) contains the majority class data, while \( D_1 \) to \( D_n \) contain limited data for the minority classes due to class imbalance (e.g., 10 per class).
 
\textbf{Memory}. The proposed method employs a dynamic queue mechanism. The queue $q_{0}$ retains the most recent instances of the majority class, with size $m$. For each minority class \(i \in \{1, \ldots, n\}\), the queue $q_{i}$ stores the most recent instances classified as class $i$, with a queue size $l$. Our setting is unsupervised, i.e., the instances are stored based on the predictions made, instead of ground truth. Due to class imbalance, $l<<m$. The queues are continuously updated with new instances assigned to the corresponding classes. A separate queue $q_{n+1}$ stores instances classified as novel. 

\subsection{Model}
\textbf{Model structure}. SCIL consists of two key parts: an AE and an MLP. The AE is employed to learn a lower-dimensional representation of the input data, capturing essential features while filtering ou redundant information. This step not only improves computational efficiency but also enhances the model’s robustness to variations in input data. The compressed representation, denoted as \( \boldsymbol{emb}^t \in \mathbb{R}^k \), serves as input to the MLP. The MLP is chosen for its efficiency in handling structured feature spaces and its ability to perform multi-class classification with nonlinear decision boundaries. By operating on the lower-dimensional embedding, the MLP benefits from reduced complexity and is less prone to overfitting. 
The entire model is defined as \( h \), i.e., \( \hat{y}^t = h(\boldsymbol{x}^t) \). The total loss \( L_{total} \) for the entire model \( h \) is shown in Eq.~(\ref{eq:total}).

\begin{equation}\label{eq:total}
\begin{split}
L_{total}(\boldsymbol{x}^1, \ldots, \boldsymbol{x}^B,
\hat{\boldsymbol{x}}^1, \ldots, \hat{\boldsymbol{x}}^B,
y^1, \ldots, y^B,
\hat{y}^1, \ldots, \hat{y}^B) \\
= \alpha \cdot L_{recon}(\boldsymbol{x}^1, \ldots, \boldsymbol{x}^B,
\hat{\boldsymbol{x}}^1, \ldots, \hat{\boldsymbol{x}}^B) \\
+ (1 - \alpha) \cdot L_{clf}(y^1, \ldots, y^B,
\hat{y}^1, \ldots, \hat{y}^B), \quad \alpha \in (0,1)
\end{split}
\end{equation}

This joint loss aims to simultaneously optimize both task performance and representation learning. The classification loss $L_{clf}$ ensures discriminative capability for known classes, while the reconstruction loss $L_{recon}$ encourages the model to preserve sufficient input information in the latent representation, improving its sensitivity to unseen classes.

\textbf{(1) AE: $\mathbb{R}^d \to \mathbb{R}^d$}. 
It is a specialized type of neural network designed to replicate its input\upcite{goodfellow2016deep}. It consists of an encoder ($\mathbb{R}^d \to \mathbb{R}^k$) followed by a decoder 
($\mathbb{R}^k \to \mathbb{R}^d$) and is trained to minimize the reconstruction loss between an input $\boldsymbol{x}^t \in \mathbb{R}^d$ and its reconstructed output $\hat{\boldsymbol{x}}^t \in \mathbb{R}^d$. The reconstruction loss can be the binary cross-entropy or the sum of squared differences. 
For a batch of instances with number $B$, both reconstruction losses can be represented as Eq.~(\ref{eq:cebatch}) and Eq.~(\ref{eq:ssbatch}).

\begin{equation}\label{eq:cebatch}
\begin{split}
&L_{recon}(\boldsymbol{x}^1, \ldots, \boldsymbol{x}^B,
\hat{\boldsymbol{x}}^1, \ldots, \hat{\boldsymbol{x}}^B)= \\
&-\sum_{j=1}^B \sum_{i=1}^d
\left[
x_{i}^j \log \hat{x}_{i}^j
+ \left(1-x_{i}^j\right)\log\left(1-\hat{x}_{i}^j\right)
\right]
\end{split}
\end{equation}

\begin{equation}\label{eq:ssbatch}
\begin{split}
&L_{recon}(\boldsymbol{x}^1, \ldots, \boldsymbol{x}^B,
\hat{\boldsymbol{x}}^1, \ldots, \hat{\boldsymbol{x}}^B) = \\
&\qquad \frac{1}{2B}\sum_{j=1}^B\sum_{i=1}^d
(\hat{x}_{i}^j - x_{i}^j)^2
\end{split}
\end{equation}

\textbf{(2) MLP: $\mathbb{R}^k \to \mathbb{R}^{n+1}$}. 
For an instance $\boldsymbol{x}^t$, after encoding by the AE, 
the embedding $\boldsymbol{emb}^t$ is used as input to the MLP for class prediction, 
i.e., $\hat{y}^t = \mathrm{MLP}(\boldsymbol{emb}^t)$. 
The MLP operates on the principle of transforming input data through multiple layers of neurons, 
each applying a linear transformation followed by a non-linear activation function. 
Given the following parameter settings: $B$, the batch size; 
$n$, the maximum class index; $\hat{y}^t$, the prediction; 
and $y^t$, the true target labels. 
The cross-entropy loss for a batch of instances, representing the classification loss 
$L_{clf}$, is shown as Eq.~(\ref{eq:batch}).

\begin{equation}\label{eq:batch}
L_{clf}(y^1, \ldots, y^B)
= -\frac{1}{B} \sum_{j=1}^B \sum_{i=0}^n
y_i^j \log \hat{y}_i^j
\end{equation}

Notably, in offline training, true labels are available for model updates. In contrast, online training lacks access to true labels, so pseudo-labels (i.e., predictions from the previous model) are used as surrogate labels to update the model. To reduce error propagation, we incorporate a correction mechanism detailed next.

\textbf{Model prediction}. The MLP in this study consists of an input layer, multiple hidden layers, and an output layer with $n+1$ classes. Given an input embedding $\boldsymbol{emb}^t \in \mathbb{R}^k$, the network processes information through successive transformations with activation functions such as ReLU or Tanh before reaching the output layer. After passing through $L$ hidden layers, the final hidden layer output $h_d^{(L)}$ is passed to the output layer to compute logits for each class:

\begin{equation}
h_i = \mathrm{f}_o \left(
\sum_{d} w_{d i}^{(L)} \cdot h_d^{(L)} + b_i^{(L)}
\right), \quad i = 0, \ldots, n
\end{equation}

\noindent , where $h_i$ is the logit corresponding to class $i$, $w_{d i}^{(L)}$ denotes the weight matrix connecting the last hidden layer to the output layer, $b_i^{(L)}$ is the output bias term, and $f_o(\cdot)$ is the output activation function applied before classification. The final class probabilities are obtained using the Softmax function:

\begin{equation}
P(i \mid \boldsymbol{emb}^t)
= \frac{\mathrm{e}^{h_i}}{\sum_{j=0}^{n} \mathrm{e}^{h_j}},
\quad i = 0, \ldots, n .
\end{equation}

\subsection{Update}\label{sec:update}

\begin{figure}[!h]
	\centering \includegraphics[scale=0.5]{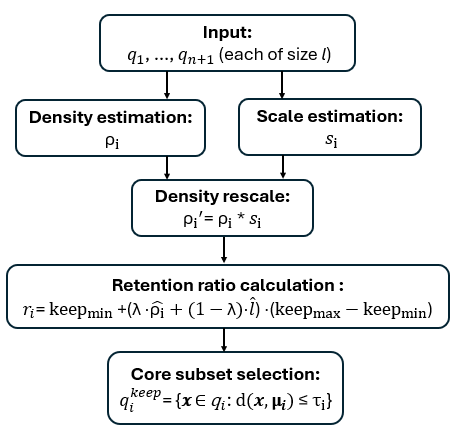}
	\caption{Flow chart of correction mechanism.}
	\label{fig:corrector_chart}
\end{figure}

\textbf{Classification Correction Mechanism}. When instances of new classes are detected, they are stored for future training. However, reconstruction-based detection may misclassify samples of seen classes as samples of new classes, causing error propagation. To mitigate this, we propose a resilient correction mechanism that adaptively purifies minority classes before resampling. By combining class density and global scale to determine retention ratios, and using the geometric median with Mahalanobis distance for robust center estimation, the mechanism filters noisy or boundary samples. This ensures oversampling is performed only on reliable core subsets, preventing noise amplification and improving robustness. Notably, as shown in Fig.~\ref{fig:SCIL}, the correction mechanism is applied only when new models are created, since frequent correction risks filtering out drifted instances and erasing historical information. The corresponding flow chart of the correction mechanism is provided in Fig.~\ref{fig:corrector_chart}.

\subsubsection*{1. Density and Scale Estimation}

Let $q_i = (\boldsymbol{x}_i^1, \boldsymbol{x}_i^2, \dots, \boldsymbol{x}_i^l)$ denote the queue of samples belonging to class $i$, where $|q_i| = l$ is the number of samples.
For each class $i$, we first estimate the local density $\rho_i$ using the inverse of the median $k$-nearest-neighbor distance:
\begin{equation} \label{eq:rho}
\rho_i = \frac{1}{\mathrm{median}_{\boldsymbol{x} \in q_i}\, \mathrm{d}_k(\boldsymbol{x}, q_i)} .
\end{equation}

To further capture the global spread of the class, we compute the class scale $s_i$ as the mean distance to a robust class center, the geometric median $\boldsymbol{\mu}_i$, which is less sensitive to outliers than the arithmetic mean:
\begin{equation} \label{eq:scale}
s_i = \frac{1}{l}\sum_{\boldsymbol{x} \in q_i}\|\boldsymbol{x} - \boldsymbol{\mu}_i\|_2 .
\end{equation}

The geometric median is defined as
\begin{equation} \label{eq:geomed}
\boldsymbol{\mu}_i = \arg \min_{\boldsymbol{m} \in \mathbb{R}^d} \sum_{\boldsymbol{x} \in q_i} \|\boldsymbol{x} - \boldsymbol{m}\|_2 ,
\end{equation}
where $\boldsymbol{m} \in \mathbb{R}^d$ denotes a candidate point in the feature space over which the minimization is performed. The optimal solution $\boldsymbol{\mu}_i$ is the point that minimizes the sum of distances and thus represents the geometric median.

\subsubsection*{2. Relative Density Compensation and Retention Ratio}

To avoid misinterpreting classes with dense cores but large overall spread, we apply relative density compensation:
$\rho'_i = \rho_i \cdot s_i$.
After normalizing $\rho'_i$ and the class size $l$ into $[0,1]$, we obtain the retention ratio $r_i$ by convex mixing:
\[
r_i = \mathrm{keep}_{\min}
+ \Big( \lambda \cdot \hat{\rho}_i + (1-\lambda)\cdot \hat{l} \Big)
\cdot \big(\mathrm{keep}_{\max}-\mathrm{keep}_{\min}\big),
\]
where $\hat{\rho}_i$ and $\hat{l}$ are the normalized density and class size, respectively, and $\lambda \in [0,1]$ balances their contributions.

\subsubsection*{3. Core Subset Selection}

From queue $q_i$, we retain the most reliable core subset of samples according to the retention ratio. 
Specifically, we keep $k_i = \max(\mathrm{min\_keep}, \lfloor r_i l \rfloor)$ samples closest to the robust center $\boldsymbol{\mu}_i$. 
The retained set is thus
\begin{equation} \label{eq:subset}
q_i^{\mathrm{keep}} = \{\boldsymbol{x} \in q_i : \mathrm{d}(\boldsymbol{x},\boldsymbol{\mu}_i) \le \tau_i\},
\end{equation}
where $\mathrm{d}(\cdot,\boldsymbol{\mu}_i)$ is the Mahalanobis distance
\begin{equation} \label{eq:mahal}
\mathrm{d}(\boldsymbol{x},\boldsymbol{\mu}_i) 
= \sqrt{(\boldsymbol{x}-\boldsymbol{\mu}_i)^{\mathrm{T}} \boldsymbol{\Sigma}_i^{-1}(\boldsymbol{x}-\boldsymbol{\mu}_i)},
\end{equation}
with $\boldsymbol{\Sigma}_i$ denoting the robust covariance matrix of class $i$, and $\tau_i$ the $k_i$-th order statistic of distances.

\textbf{Oversampling with SMOTE}. 
To address the class imbalance between the majority class and minority classes, Synthetic Minority Over-sampling Technique (SMOTE)\upcite{chawla2002smote} generates synthetic samples by interpolating between a minority instance 
$\boldsymbol{x}^a_i$ (where $i \in \{1,2,\ldots,n\}$ and $a \in \{1,2,\ldots,l\}$) 
and one of its $K$-nearest neighbors. 
Using KNN to identify $K$ closest neighbors $\in q_i$ based on Euclidean distance, 
SMOTE selects a random neighbor $\boldsymbol{x}^b_i$ and generates a synthetic sample along the line segment connecting them by scaling the difference vector by a random value in $[0,1]$. 
This process is shown in Eq.~\ref{eq:smote}, corresponding to Lines 20 and 22 of Algorithm 1.

\begin{equation}\label{eq:smote}
\boldsymbol{x}^{\mathrm{new}}_{i}
=
\boldsymbol{x}^a_{i}
+
\left(\boldsymbol{x}^b_{i}-\boldsymbol{x}^a_{i}\right)
\times \mathrm{rand}(0,1)
\end{equation}

\textbf{Training}. 
As outlined in Algorithm 1, the training process comprises three stages: 
\textbf{pretraining} (Line 2), 
\textbf{incremental training} (Line 20), 
and \textbf{new model training} upon detecting a novel class (Line 18). 
During \textbf{pretraining}, the initial model is trained on known classes $D_0, D_1, \dots, D_n$, 
and novelty thresholds $\theta_0, \theta_1, \dots, \theta_n$ are computed based on 
Eq.~(\ref{eq:threshold_0}) and Eq.~(\ref{eq:threshold_i}). 
The majority class threshold $\theta_0$ is set as the maximum reconstruction loss within a queue $q_0$ of size $m$ 
(Eq.~(\ref{eq:threshold_0})), 
while minority class thresholds are defined in Eq.~(\ref{eq:threshold_i}) with a queue size of $l$. 
For \textbf{incremental training}, the model continuously adapts to concept drift. 
If the current time is $t + \Delta$ and the last training time is $t$, retraining occurs when $\Delta = t_{\text{train}}$, 
refining the model as 
$h^{t+\Delta} = h^{t}.\mathrm{train}(\mathrm{SMOTE}(q_0, q_1, \dots, q_n))$, 
where $t_{train}$ defines the incremental training interval. 
During training, the model parameters $\boldsymbol{\Phi}$ are updated via gradient descent:
\[
\boldsymbol{\Phi}^{t+\Delta} 
= 
\boldsymbol{\Phi}^{t} 
- 
\eta \nabla_{\boldsymbol{\Phi}} \mathcal{L}_{\text{total}}(h^{t}, q_0, q_1, \dots, q_n),
\]
where $\eta$ is the learning rate and 
$\nabla_{\boldsymbol{\Phi}} \mathcal{L}_{\text{total}}$ is the gradient (partial derivative) of the loss function 
$\mathcal{L}_{\text{total}}$ with respect to the model parameters $\boldsymbol{\Phi}$ at time step $t$. 
At each step, thresholds $\theta_0, \theta_1, \dots, \theta_n$ are updated (Lines 20–21). 
If the new class queue $q_{n+1}$ reaches size $l$ (Line 15), it is recognized as a new class, triggering 
\textbf{new model training} (Lines 16–18). 
This process continues as incremental classes emerge.

\begin{equation}\label{eq:threshold_0}
\theta_{0}
=
\max\!\Big(
L_{\text{recon}}\big(
(\boldsymbol{x}_{0}^1, \hat{\boldsymbol{x}}_{0}^1),\ldots,
(\boldsymbol{x}_{0}^{m},\hat{\boldsymbol{x}}_{0}^{m})
\big)
\Big)
\end{equation}

\begin{equation}\label{eq:threshold_i}
\theta_{i}
=
\max\!\Big(
L_{\text{recon}}\big(
(\boldsymbol{x}_{i}^1, \hat{\boldsymbol{x}}_{i}^1),\ldots,
(\boldsymbol{x}_{i}^{l},\hat{\boldsymbol{x}}_{i}^{l})
\big)
\Big)
\end{equation}

Notably, when a new class appears, we train a new model using queues that contain recent examples from all classes. 
This allows the model to revisit earlier patterns during training and helps mitigate catastrophic forgetting by preserving previously learned representations.

\subsection{Computational Analysis}\label{sec:com_ana}

\textbf{Memory requirements:} 
The proposed method uses $n + 2$ queues: $n$ for known minority classes, one for the majority class, and one for the novel class. 
A queue of size $m$ stores majority class instances, while the remaining $n + 1$ queues of size $l$ hold samples from minority and novel classes. 
The space complexity is thus $\mathcal{O}\!\left(m + (n + 1)\times l\right)$, where $m$ and $l$ are fixed and $l$ is small (e.g., $10$). 
Importantly, SCIL satisfies one of the most desired properties of learning in nonstationary environments, which is having a fixed amount of memory for any storage\upcite{gama2014survey}. 
Only when a new class is detected does the total number of queues increase by one, incrementing memory by $\mathcal{O}(l)$. 
In addition to data storage, the neural network requires memory for its weights, proportional to the number of model parameters.

\textbf{Prediction Stage:} 
The prediction stage involves both classification and new class detection. 
The input is first processed by the AE to compute the reconstruction loss, followed by MLP-based classification using the AE’s latent embedding. 
Finally, the reconstruction loss is compared with the threshold of the predicted class for new class detection.


\textbf{Training Stage:} 
During the training stage, SCIL performs incremental learning at regular intervals $t_{\text{train}}$ (e.g., every $2000$ steps), using a small number of epochs per session (e.g., $10$). 
At each session, a new threshold for each class is computed based on queues $q_0, q_1, \ldots, q_n$. 
The time complexity of training is $\mathcal{O}((n+1)\times m)$. 
Whenever $q_{n+1}$ reaches capacity, all $n+1$ minority-class queues are used to train a new model, and class thresholds are updated. 
In this case, the correction mechanism and SMOTE oversampling are applied to queues $q_1, \ldots, q_{n+1}$, ensuring efficient memory usage during online operation. 
Overall, four operations are considered: correction, SMOTE, training, and threshold calculation. 
Since correction and SMOTE focus only on minority classes, their cost is ignored here. 
The training cost, dominated by the $n+1$ queues of size $m$, has a time complexity of $\mathcal{O}((n+1)\times m)$. 
Therefore, the time complexity grows with the number of minority classes $n$. 
Note that if a new class appears infrequently (e.g., a rare anomaly), it will not significantly increase the training cost, since threshold updates and model retraining are only triggered when the corresponding queue reaches capacity. 
Moreover, rarely appearing minority classes can be forgotten after long periods of inactivity, ensuring that the overall time complexity remains bounded.

\section{Experimental Setup}\label{sec:exp_setup}

\subsection{Datasets} 
\begin{figure}[!t]
\begin{subfigure}{.5\columnwidth}
  \centering
\includegraphics[width=1\columnwidth]{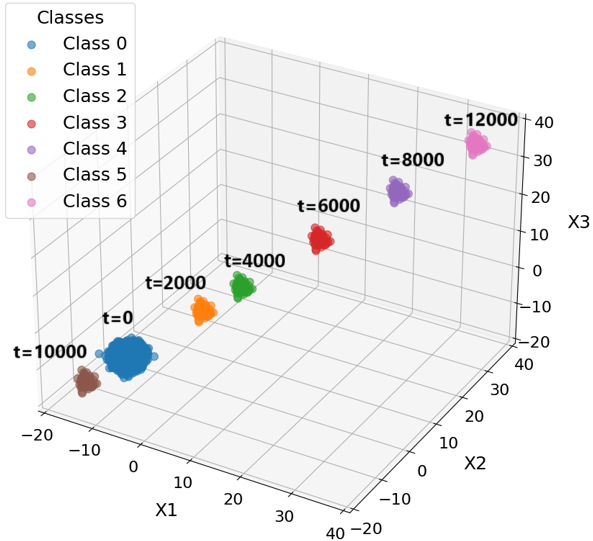}
  \caption{Blob at t=15000}
\label{fig:blob}
\end{subfigure}%
\begin{subfigure}{.5\columnwidth}
  \centering
\includegraphics[width=1\columnwidth]{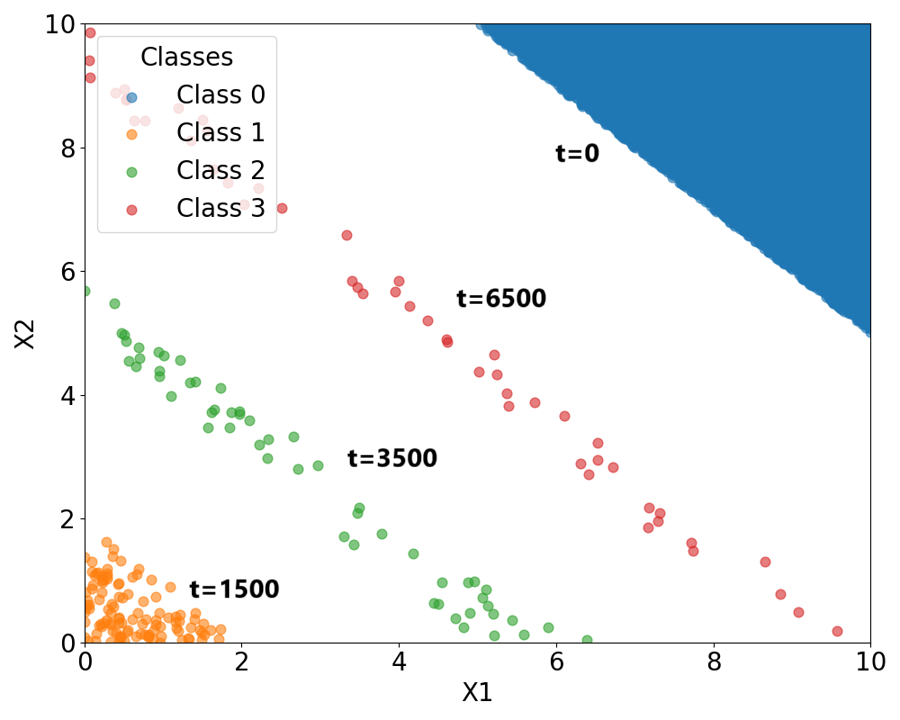}
  \caption{Sea at t=15000}
\label{fig:sea}
\end{subfigure}
\centering
\caption{Screenshots of datasets Blob and Sea (Timestamps mark each class's first appearance).}
\label{fig:screenshot}
\end{figure}

We first describe the datasets, which vary in features, imbalance rates, and class counts (Table~\ref{tab:gen_data}). They incorporate sudden, incremental, and recurrent drifts in varied combinations (Table~\ref{tab:drift}) to simulate realistic scenarios. Fig.~\ref{fig:screenshot} shows Blob and Sea dataset snapshots at $t=15000$.

\subsubsection{Description} 
Our experiments consider both synthetic (Sea\upcite{street2001streaming}, Blob\upcite{malialis2022nonstationary}, Vib\upcite{li2024unsupervised}) and real-world (WDN, MNIST\upcite{lecun1998gradient}, KDD99\upcite{Lichman2013}, Forest\upcite{blackard1999comparative}, Sensorless\upcite{dataset_for_sensorless_drive_diagnosis_325}, Shuttle\upcite{dua2017uci}) datasets.

\textbf{Sea:} its class boundary is set by inequations, as shown in Table~\ref{tab:drift} in the `Before Drift' column.

\textbf{Blob:} each class is an isotropic Gaussian blob and noise exists due to the standard deviation of the blobs. The centres and standard deviations are shown in the `Before Drift' column. 

\textbf{Vib:} a 10-dimensional dataset consisting of simulated equipment vibration data in industrial manufacturing.

\textbf{WDN:} Water Distribution Network data from chlorine sensors (single node). Contamination events at different nodes form minority classes. Simulated using WNTR\cite{klise2017software} on Hanoi WDN\cite{fujiwara1990two} (detection at node 9). Six months of 30-minute samples (8640 points). Concept drift: sensor offset randomly scaled 1.3–1.5×.

\textbf{MNIST:} a benchmark dataset of handwritten digits, challenging streaming methods with its 784-dimensional features. Data stream is formed with selected digits shown in Table~\ref{tab:drift}.

\textbf{Forest:} it includes U.S. Forest Service cartographic data for predicting cover type in 30×30m cells from Roosevelt National Forest, Colorado.

\textbf{KDD99:} a benchmark dataset from DARPA and MIT Lincoln Labs (1999), simulates diverse intrusions in a military network for intrusion detection.

\textbf{Sensorless:} it contains electric drive signals from 11 conditions (varying speeds/loads/forces), captured via current probes/oscilloscope, covering intact and defective components.

\textbf{Shuttle:} it contains NASA Space Shuttle flight control system process records for classifying propulsion system operating states.

\subsubsection{Drift and incremental classes}
As Table~\ref{tab:gen_data} shows, varying initial class counts ensure pretraining diversity: 7 classes for Blob and 10 for MNIST demonstrate new class detection and incremental handling. Table~\ref{tab:drift}'s 'Drifts Time' column indicates when drifts occur, with various drift type combinations across datasets evaluating robustness (see 'Drift Type' column). 'Before Drift'/'After Drift' descriptions highlight class changes. For example, in the Blob dataset, a drift occurs at time 7000. Before this point, four classes are present, but after the drift, new classes (C4, C5, and C6) appear, and the center of C0 shifts to [-9.5, -9.5, -9.5]. For the Shuttle dataset, the presence and nature of drift remain unknown.

\begin{table}[!t]
\caption{General description of datasets.}\label{tab:gen_data}
\begin{adjustbox}
{width=0.5\textwidth}
\begin{tabular}{|c|c|c|c|c|c|c|}
\hline
\textbf{Type} & \textbf{Dataset} & \textbf{\#Features} & \textbf{\#Data} & \textbf{\begin{tabular}[c]{@{}c@{}}Initial\\ \#Classes\end{tabular}} & \textbf{\begin{tabular}[c]{@{}c@{}}Total\\ \#Classes\end{tabular}} & \textbf{\begin{tabular}[c]{@{}c@{}}Imbalance\\ Rate\end{tabular}} \\ \hline
\multirow{3}{*}{Synthetic}  
& Sea        & 2   & 15000  & 2 & 4 & 1.3\% \\ \cline{2-7} 
& Vib        & 10  & 15000  & 2 & 4 & 2.0\% \\ \cline{2-7} 
& Blob       & 3   & 15000  & 2 & 7 & 2.0\% \\ \hline
\multirow{5}{*}{Real-world} 
& WDN        & 1   & 8640   & 1 & 3 & 1.2\% \\ \cline{2-7} 
& MNIST      & 784 & 10000  & 2 & 10 & 4.5\% \\ \cline{2-7} 
& KDD99      & 116 & 15000  & 2 & 4 & 2.4\% \\ \cline{2-7} 
& Forest     & 52  & 15000  & 2 & 4 & 2.4\% \\ \cline{2-7} 
& Sensorless & 48  & 5000   & 2 & 3 & 4.0\% \\ \cline{2-7} 

& Shuttle &9  & 55827   & 2 & 4 & 21.9\% \\ \hline
\end{tabular}
\end{adjustbox}
\end{table}

\begin{table}[!t]
\caption{Drift description of datasets.}\label{tab:drift}
\begin{adjustbox}{width=0.5\textwidth}
\begin{tabular}{|c|c|c|c|c|}
\hline
\textbf{Dataset} & \textbf{Drifts Time} & \textbf{Drift Type} & \textbf{Before Drift} & \textbf{After Drift} \\ \hline
Sea & \begin{tabular}[c]{@{}c@{}}5000\\ 10000\end{tabular} & Recurrent & \begin{tabular}[c]{@{}c@{}}C0: $x_1 + x_2 > 15$\\ C1: $x_1 + x_2 \leq 2$\\ C2: $5 < x_1 + x_2 < 6$\\ C3: $9 < x_1 + x_2 < 10$\\ $x_1, x_2 \in [0,1]$\end{tabular} & \begin{tabular}[c]{@{}c@{}}C0: $x_1 + x_2 > 16$\\ C1: $x_1 + x_2 \leq 1.5$\\ C2: $5 < x_1 + x_2 < 6.5$\\ $x_1, x_2 \in [0,1]$\end{tabular} \\ \hline 
Vib & \begin{tabular}[c]{@{}c@{}}5000-5050\\ 10000-10050\end{tabular} & \begin{tabular}[c]{@{}c@{}}Recurrent\\ Incremental\end{tabular} & \begin{tabular}[c]{@{}c@{}}C0: mean=0, std=1\\ C1: mean=5, std=1\\ C2: mean=10, std=1\\ C3: mean=20, std=1\end{tabular} & \begin{tabular}[c]{@{}c@{}}C0: mean=0.5, std=1\\ C1: mean=5.5, std=1\\ C2: mean=10.5, std=1\end{tabular} \\ \hline 
Blob & 7000 & Abrupt & \begin{tabular}[c]{@{}c@{}}C0: center=[-10,-10,-10]\\ C1: center=[0,0,0]\\ C2: center=[5,5,5]\\ C3: center=[15,15,15]\\ std=1.0\end{tabular} & \begin{tabular}[c]{@{}c@{}}C0: center=[-9.5,-9.5,-9.5]\\ C4: center=[25,25,25]\\ C5: center=[-15,-15,-15]\\ C6: center=[35,35,35]\\ std=1.0\end{tabular} \\ \hline
WDN & 4000 & Abrupt & \begin{tabular}[c]{@{}c@{}}C0: chlorine=0.2\\ C1: contamination at N6\end{tabular} & \begin{tabular}[c]{@{}c@{}}C0: chlorine=0.2\\ C2: contamination at N9\\ with sensor offset at N11\end{tabular} \\ \hline 
MNIST & 5000 & Abrupt & \begin{tabular}[c]{@{}c@{}}C0: digit 0\\ C1: digit 1\\ C2: digit 2\\ C3: digit 3\\ C4: digit 4\end{tabular} & \begin{tabular}[c]{@{}c@{}}C0: digit 0\\ C5: digit 5\\ C6: digit 6\\ C7: digit 7\\ C8: digit 8\\ C9: digit 9\\ shift +/-1\%\end{tabular} \\ \hline 
KDD99 & \begin{tabular}[c]{@{}c@{}}5000\\ 10000\end{tabular} & Recurrent & \begin{tabular}[c]{@{}c@{}}C0: normal\\ C1: smurf\\ C2: neptune\\ C3: satan\end{tabular} & \begin{tabular}[c]{@{}c@{}}C0: normal\\ C1: smurf\\ C2: neptune\\ C3: satan\\ values *0.9\end{tabular} \\ \hline
Forest & \begin{tabular}[c]{@{}c@{}}5000-5050\\ 10000\end{tabular} & \begin{tabular}[c]{@{}c@{}}Incremental\\ Abrupt\end{tabular} & \begin{tabular}[c]{@{}c@{}}C0: covertype1\\ C1: covertype3\\ C2: covertype7\\ C3: covertype4\end{tabular} & \begin{tabular}[c]{@{}c@{}}C0: covertype1\\ C1: covertype3\\ C2: covertype7\\ C3: covertype4\\ values *0.9\end{tabular} \\ \hline 
Sensorless & 2500-2600 & Gradual & \begin{tabular}[c]{@{}c@{}}C0: class 1\\ C1: class 2\end{tabular} & \begin{tabular}[c]{@{}c@{}}C0: class 1\\ C2: class 11\\ value *0.95\end{tabular} \\ \hline

Shuttle & Unknown & Unknown & \begin{tabular}[c]{@{}c@{}}Unknown\end{tabular} & \begin{tabular}[c]{@{}c@{}}Unknown\end{tabular} \\ \hline

\end{tabular}
\end{adjustbox}
\end{table}

\begin{table*}[!t]
\caption{Description of methods.}\label{tab:method_des}
\begin{adjustbox}{width=1.0\textwidth}
 \centering
\begin{tabular}{|c|c|c|c|c|}
\hline
\textbf{Method} & \textbf{Learning  Category}      & \textbf{Classification Mechanism} & \textbf{New Class Detection Mechanism}    & \textbf{Model Update Mechanism}                       \\ \hline
Baseline        & Representation Learning                    & AE+MLP                            & Reconstruction loss threshold-based & Update with Buffered New Class                        \\ \hline
iForest+MULTI   & Tree-based                 & iForest+Distance-based            & Distance Threshold-based            & Update with Buffered New Class                        \\ \hline
LOF+MULTI       & Outlier Detection                 & LOF+Distance-based                & Distance Threshold-based            & Update with Buffered New Class                        \\ \hline
MINAS           & Online Clustering                & Cluster-based                     & Distance Threshold-based            & Update with Buffered New Class                        \\ \hline
CPOCEDS           & Online Clustering                & Cluster-based                     & Distance Threshold-based            & Incremental Clustering with Concept Preservation         \\ \hline
SNDProb         & Probabilistic Streaming Learning & GMM+Probabilistic-based           & Probability Threshold-based         & Update Probability Distribution with EM               \\ \hline

KNNENS          & Ensemble Cluster Learning        & KNN Ensemble                      & KNN Ensemble-based                  & Update with Buffered New Class                        \\ \hline

OCGCD          & Representation Learning + Clustering        & Cluster-based                      & Distance-based Novelty Detection                  & Adaptive Cluster Growing and Merging                   \\ \hline

UDOR          & Representation Learning + Clustering        & Classifier + Cluster-based                      &  Unknown-score / Distance-based                  & Novel-class Discovery via Unsupervised Clustering                        \\ \hline

SCIL (Ours)     & Representation Learning                    & AE+MLP                            & Reconstruction loss threshold-based & Incremental Learning+Update with Buffered New Class   \\ \hline
\end{tabular}
\end{adjustbox}
\end{table*}

\begin{table*}[!t]
\centering
\caption{Hyper-parameter values for SCIL.}\label{tab:param}
\begin{adjustbox}{width=1\textwidth}
\begin{tabular}{|c|c|ccc|cccc|cccclcc|}
\hline
                                     &            & \multicolumn{3}{c|}{\textbf{AE}}                                                                                                                                                                                                          & \multicolumn{4}{c|}{\textbf{MLP}}                                                                                                                                                                                                                                                                                                                 & \multicolumn{7}{c|}{\textbf{Common}}                                                                                                                                                                                                                                                                                                                                                                                                                                                                                                           \\ \hline
                                     &            & \multicolumn{1}{c|}{\begin{tabular}[c]{@{}c@{}}Hidden\\ Layers\end{tabular}} & \multicolumn{1}{c|}{\begin{tabular}[c]{@{}c@{}}Output\\ Activition\end{tabular}} & \begin{tabular}[c]{@{}c@{}}Reconstruction\\ Loss\end{tabular}           & \multicolumn{1}{c|}{\begin{tabular}[c]{@{}c@{}}Input\\ Dimension\end{tabular}} & \multicolumn{1}{c|}{\begin{tabular}[c]{@{}c@{}}Output\\ Dimension\end{tabular}}                 & \multicolumn{1}{c|}{\begin{tabular}[c]{@{}c@{}}Hidden\\ Layers\end{tabular}} & \begin{tabular}[c]{@{}c@{}}Classification\\ Loss\end{tabular}                   & \multicolumn{1}{c|}{Optimizer}             & \multicolumn{1}{c|}{\begin{tabular}[c]{@{}c@{}}Learning\\ Rate\end{tabular}} & \multicolumn{1}{c|}{\begin{tabular}[c]{@{}c@{}}Batch\\ Size\end{tabular}} & \multicolumn{1}{c|}{\begin{tabular}[c]{@{}c@{}}Number of\\ Epochs (offline)\end{tabular}} & \multicolumn{1}{c|}{\begin{tabular}[c]{@{}c@{}}Number of\\ Epochs (online)\end{tabular}} & \multicolumn{1}{c|}{\begin{tabular}[c]{@{}c@{}}Hidden\\ Activation\end{tabular}} & \begin{tabular}[c]{@{}c@{}}Weight\\ Initializer\end{tabular} \\ \hline
\multirow{4}{*}{\textbf{Synthetic}}  & Sea        & \multicolumn{1}{c|}{8}                                                       & \multicolumn{1}{c|}{\multirow{9}{*}{Sigmoid}}                                    & \multirow{9}{*}{\begin{tabular}[c]{@{}c@{}}Square\\ Error\end{tabular}} & \multicolumn{1}{c|}{2}                                                         & \multicolumn{1}{c|}{\multirow{9}{*}{\begin{tabular}[c]{@{}c@{}}Initial\\ \#Class\end{tabular}}} & \multicolumn{1}{c|}{2}                                                       & \multirow{9}{*}{\begin{tabular}[c]{@{}c@{}}Binary\\ Cross-entropy\end{tabular}} & \multicolumn{1}{c|}{\multirow{9}{*}{Adam}} & \multicolumn{1}{c|}{\multirow{9}{*}{0.001}}                                  & \multicolumn{1}{c|}{\multirow{9}{*}{32}}                                  & \multicolumn{1}{c|}{\multirow{9}{*}{20}}                                                  & \multicolumn{1}{l|}{\multirow{9}{*}{10}}                                                 & \multicolumn{1}{c|}{\multirow{9}{*}{LeakyReLU}}                                  & \multirow{9}{*}{He\_Normal}                                  \\ \cline{2-3} \cline{6-6} \cline{8-8}
                                     & Vib        & \multicolumn{1}{c|}{8}                                                       & \multicolumn{1}{c|}{}                                                            &                                                                         & \multicolumn{1}{c|}{2}                                                         & \multicolumn{1}{c|}{}                                                                           & \multicolumn{1}{c|}{2}                                                       &                                                                                 & \multicolumn{1}{c|}{}                      & \multicolumn{1}{c|}{}                                                        & \multicolumn{1}{c|}{}                                                     & \multicolumn{1}{c|}{}                                                                     & \multicolumn{1}{l|}{}                                                                    & \multicolumn{1}{c|}{}                                                            &                                                              \\ \cline{2-3} \cline{6-6} \cline{8-8}
                                     & Blob       & \multicolumn{1}{c|}{8}                                                       & \multicolumn{1}{c|}{}                                                            &                                                                         & \multicolumn{1}{c|}{2}                                                         & \multicolumn{1}{c|}{}                                                                           & \multicolumn{1}{c|}{2}                                                       &                                                                                 & \multicolumn{1}{c|}{}                      & \multicolumn{1}{c|}{}                                                        & \multicolumn{1}{c|}{}                                                     & \multicolumn{1}{c|}{}                                                                     & \multicolumn{1}{l|}{}                                                                    & \multicolumn{1}{c|}{}                                                            &                                                              \\ \cline{2-3} \cline{6-6} \cline{8-8}
                                     & WDN        & \multicolumn{1}{c|}{2}                                                       & \multicolumn{1}{c|}{}                                                            &                                                                         & \multicolumn{1}{c|}{2}                                                         & \multicolumn{1}{c|}{}                                                                           & \multicolumn{1}{c|}{2}                                                       &                                                                                 & \multicolumn{1}{c|}{}                      & \multicolumn{1}{c|}{}                                                        & \multicolumn{1}{c|}{}                                                     & \multicolumn{1}{c|}{}                                                                     & \multicolumn{1}{l|}{}                                                                    & \multicolumn{1}{c|}{}                                                            &                                                              \\ \cline{1-3} \cline{6-6} \cline{8-8}
\multirow{5}{*}{\textbf{Real-world}} & MNIST      & \multicolumn{1}{c|}{[512,256,64,32]}                                         & \multicolumn{1}{c|}{}                                                            &                                                                         & \multicolumn{1}{c|}{20}                                                        & \multicolumn{1}{c|}{}                                                                           & \multicolumn{1}{c|}{[16,8]}                                                  &                                                                                 & \multicolumn{1}{c|}{}                      & \multicolumn{1}{c|}{}                                                        & \multicolumn{1}{c|}{}                                                     & \multicolumn{1}{c|}{}                                                                     & \multicolumn{1}{l|}{}                                                                    & \multicolumn{1}{c|}{}                                                            &                                                              \\ \cline{2-3} \cline{6-6} \cline{8-8}
                                     & KDD99      & \multicolumn{1}{c|}{64}                                                      & \multicolumn{1}{c|}{}                                                            &                                                                         & \multicolumn{1}{c|}{20}                                                        & \multicolumn{1}{c|}{}                                                                           & \multicolumn{1}{c|}{[16,8]}                                                  &                                                                                 & \multicolumn{1}{c|}{}                      & \multicolumn{1}{c|}{}                                                        & \multicolumn{1}{c|}{}                                                     & \multicolumn{1}{c|}{}                                                                     & \multicolumn{1}{l|}{}                                                                    & \multicolumn{1}{c|}{}                                                            &                                                              \\ \cline{2-3} \cline{6-6} \cline{8-8}
                                     & Forest     & \multicolumn{1}{c|}{32}                                                      & \multicolumn{1}{c|}{}                                                            &                                                                         & \multicolumn{1}{c|}{20}                                                        & \multicolumn{1}{c|}{}                                                                           & \multicolumn{1}{c|}{[16,8]}                                                  &                                                                                 & \multicolumn{1}{c|}{}                      & \multicolumn{1}{c|}{}                                                        & \multicolumn{1}{c|}{}                                                     & \multicolumn{1}{c|}{}                                                                     & \multicolumn{1}{l|}{}                                                                    & \multicolumn{1}{c|}{}                                                            &                                                              \\ \cline{2-3} \cline{6-6} \cline{8-8}
                                     & Sensorless & \multicolumn{1}{c|}{32}                                                      & \multicolumn{1}{c|}{}                                                            &                                                                         & \multicolumn{1}{c|}{20}                                                        & \multicolumn{1}{c|}{}                                                                           & \multicolumn{1}{c|}{[16,8]}                                                  &                                                                                 & \multicolumn{1}{c|}{}                      & \multicolumn{1}{c|}{}                                                        & \multicolumn{1}{c|}{}                                                     & \multicolumn{1}{c|}{}                                                                     & \multicolumn{1}{l|}{}                                                                    & \multicolumn{1}{c|}{}                                                            &                                                              \\ \cline{2-3} \cline{6-6} \cline{8-8}
                                     & Shuttle    & \multicolumn{1}{c|}{8}                                                       & \multicolumn{1}{c|}{}                                                            &                                                                         & \multicolumn{1}{c|}{2}                                                         & \multicolumn{1}{c|}{}                                                                           & \multicolumn{1}{c|}{2}                                                       &                                                                                 & \multicolumn{1}{c|}{}                      & \multicolumn{1}{c|}{}                                                        & \multicolumn{1}{c|}{}                                                     & \multicolumn{1}{c|}{}                                                                     & \multicolumn{1}{l|}{}                                                                    & \multicolumn{1}{c|}{}                                                            &                                                              \\ \hline
\end{tabular}
\end{adjustbox}
\end{table*}

\subsection{Methods}

Our comparative study includes a diverse set of methods. Table~\ref{tab:method_des} summarizes their key characteristics. `Update with Buffered New Class' indicates the model is updated once the buffer of new-class data is full.

\textbf{Baseline}: In this approach, we begin the pre-training process with labeled data. The training phase is conducted offline using 1000 majority examples and 30 instances from each minority class. This pre-trained model corresponds to the initial phase of SCIL and is equivalent to running SCIL without incremental learning.

\textbf{SCIL}: This is our proposed method. Details are provided in Sec.~\ref{sec:method}.

\textbf{iForest+MULTI}\upcite{liu2008isolation}: iForest (see Sec.~\ref{sec:anodetec}) is extended to multi-class classification by combining binary anomaly detection (treating majority as normal, minority as anomalous) with Euclidean distance comparison. For each class $i$, if an instance's distance to the class center is below a threshold (the maximum distance among known instances of that class), it is assigned to class $i$; otherwise, it is flagged as a novel class. For fairness, SCIL's incremental class mechanism is incorporated.

\textbf{LOF+MULTI}\upcite{breunig2000lof}: For comparison, we use LOF\upcite{breunig2000lof}, as described in Sec.~\ref{sec:anodetec}. Similar to iForest+MULTI, instances are first classified as normal or anomalous, and then the same multi-classification and new class detection mechanism is applied. SCIL’s incremental class mechanism is included. 

\textbf{MINAS}\upcite{de2016minas}: A state-of-the-art online clustering method described in Sec.~\ref{sec:mulcla}.

\textbf{CPOCEDS}\upcite{jafseer2024cpoceds}: A state-of-the-art concept preserving online clustering method described in Sec.~\ref{sec:mulcla}.

\textbf{SNDProb}\upcite{carreno2022sndprob}: A probabilistic streaming novelty detection method. See Sec.~\ref{sec:mulcla} for details.

\textbf{KNNENS}\upcite{zhang2022knnens}: A cutting-edge classification method based on a K-Nearest Neighbors ensemble. See Sec.~\ref{sec:mulcla} for details.

\textbf{OCGCD}\upcite{park2024online}: A state-of-the-art online clustering method described in Sec.~\ref{sec:mulcla}.

\textbf{UDOR}\upcite{alfarisy2025towards}: A state-of-the-art online clustering method described in Sec.~\ref{sec:mulcla}.

\subsection{Performance metrics}

To evaluate the predictive accuracy of algorithms over time, \cite{mu2017classification} proposes the metric EN\_Accuracy, which we adopt to compute dynamically at each time step $t$. 
Let $N(t)$ be the number of processed instances up to time $t$, 
$A_n(t)$ the number of correctly identified instances of emerging classes, 
and $A_o(t)$ the number of correctly classified instances of known classes at time $t$. 
The time-dependent EN Accuracy is defined as follows:

\begin{equation}\label{eq:en_acc}
\mathrm{EN\_Accuracy}(t)
=
\frac{A_n(t) + A_o(t)}{N(t)} ,
\end{equation}
where $N(t) = t$ in an online learning setting where one instance arrives at each time step.

The overall accuracy is usually used to evaluate classifiers. 
However, in the presence of class imbalance, this evaluation method becomes biased towards the majority classes. 
To avoid this issue, we adopt a widely accepted metric that is insensitive to class imbalance: geometric mean (G-mean)\upcite{sun2006boosting}, defined as:
\begin{equation}\label{eq:gmean}
\mathrm{G\text{-}mean}
=
\sqrt[n]{\prod_{i=0}^{n} R_i}
\end{equation}

\noindent where $R_i$ is the recall of class $i$ and $i \in \{0, 1, \ldots, n\}$. 
G-mean is not only insensitive to class imbalance, but it also has some important properties as it is high when all recalls are high and when their difference is small\upcite{he2008learning}. 
We adopt prequential evaluation with a fading factor of $0.99$, which avoids the need for a holdout set and converges to Bayes error on stationary data~\upcite{gama2013evaluating}. 
G-mean is computed at each time step and averaged over $10$ runs, with error bars showing standard error.

Overall, EN\_Accuracy reflects overall performance across known and new classes, while G-mean evaluates classification performance in the presence of class imbalance and tracks performance as class distributions evolve.

\section{Experimental Results}\label{sec:exp_results}

The hyperparameters for SCIL are shown in Table~\ref{tab:param}. The full list will be provided in our released code. We set the queue sizes $m=1000$, $l=30$, and the training interval $t_{train}=2000$ for all datasets. The DBSCAN parameters are set to $\epsilon=0.5$ and $P=5$. This study focuses on one majority class and multiple minority classes, simulating scenarios with multiple anomalies.

\subsection{Ablation studies}
In the ablation study, we first investigate the need for using embeddings as input to the MLP rather than using raw instances directly. Then, we experiment with various weights for the $L_{recon}$ and $L_{clf}$ within the $L_{total}$ to identify the optimal combination. Then, we investigate the impact of the oversampling technique on model performance, highlighting its effectiveness in addressing data imbalance. Finally, we evaluate the contribution of the correction mechanism to the overall classification performance and the sensitivity of our method to class imbalance.

 
 

\begin{figure*}[!t]
 \begin{subfigure}{.33\textwidth} 
  \centering
  \includegraphics[width=0.98\textwidth]{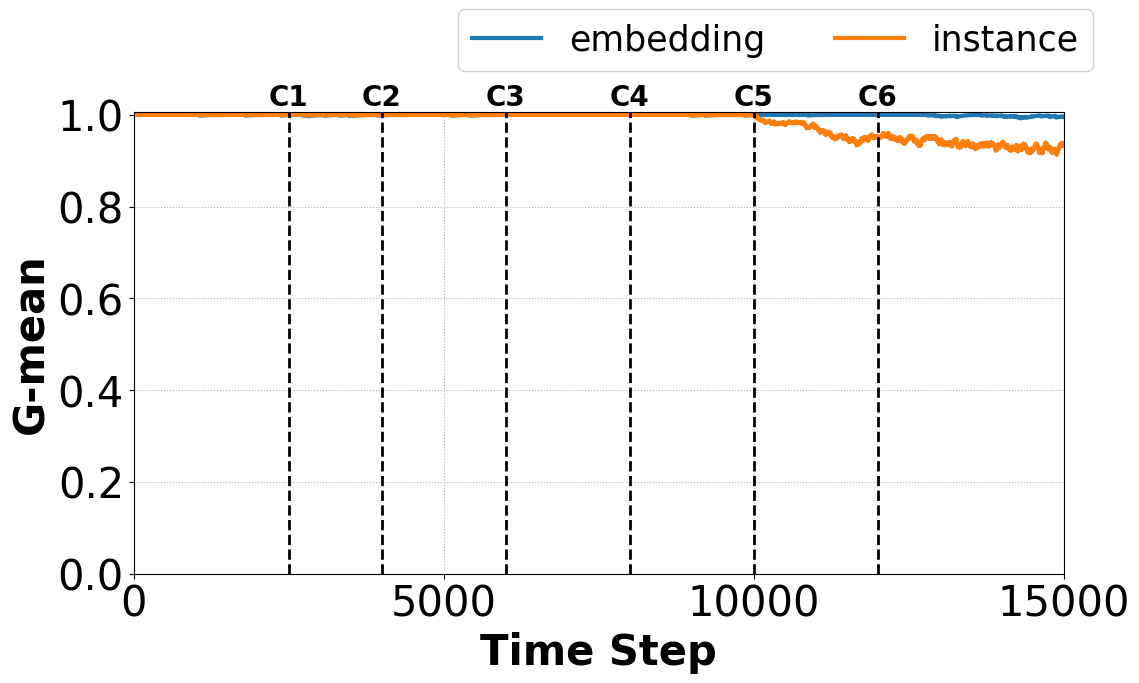} 
  \caption{Blob}
  \label{fig:blob_emb_abla}
 \end{subfigure}%
 \begin{subfigure}{.33\textwidth}
  \centering
  \includegraphics[width=0.98\textwidth]{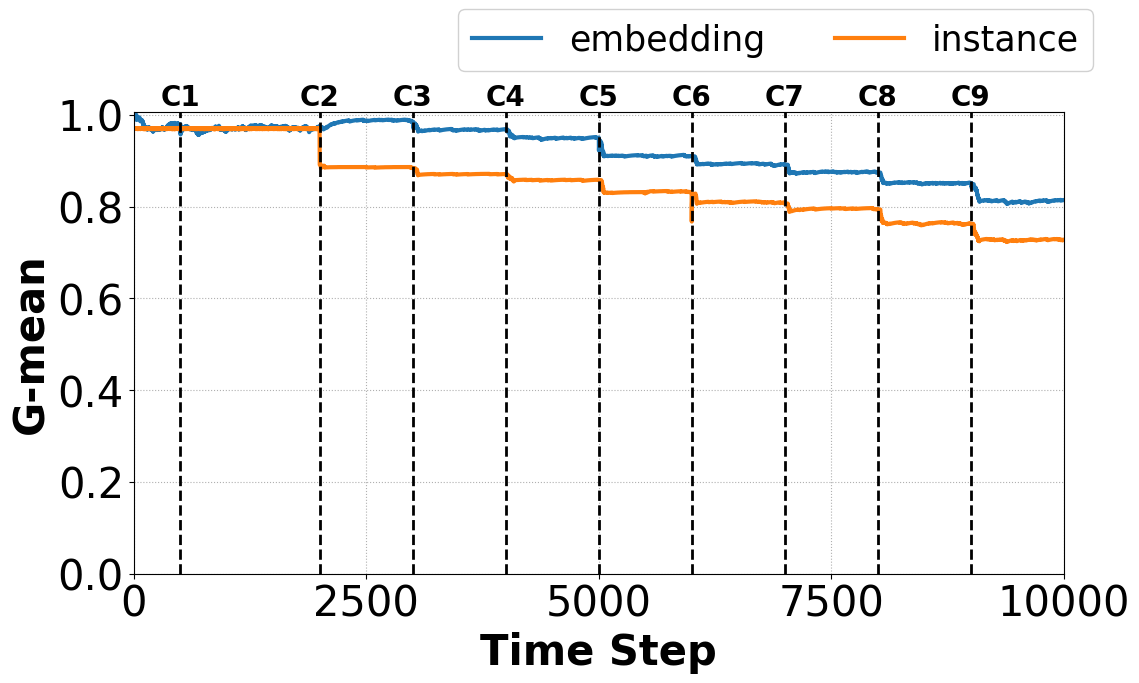} 
  \caption{MNIST}
  \label{fig:mnist_emb_abla}
 \end{subfigure}%
 \begin{subfigure}{.33\textwidth}
  \centering
  \includegraphics[width=0.98\textwidth]{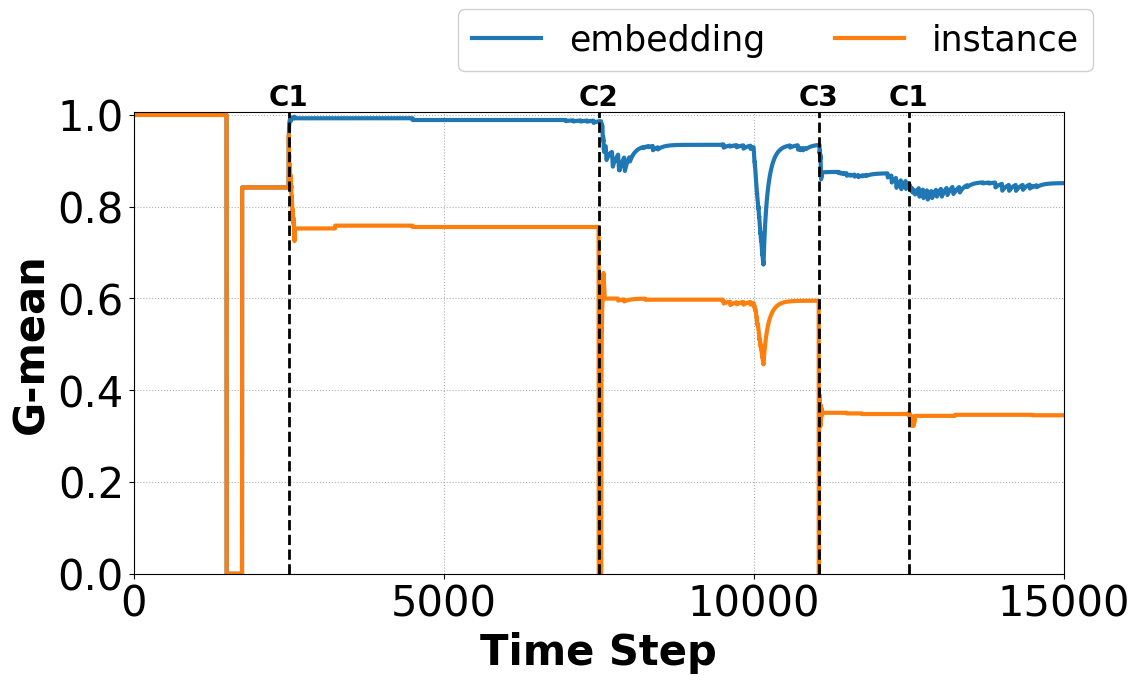} 
  \caption{Forest}
  \label{fig:forest_emb_abla}
 \end{subfigure}%
\caption{Comparison of different input configurations in nonstationary environments.}
\label{fig:emb_compare}
\end{figure*}

\begin{figure*}[!t]
    \begin{subfigure}{.33\textwidth} 
      \centering
      \includegraphics[width=0.98\textwidth]{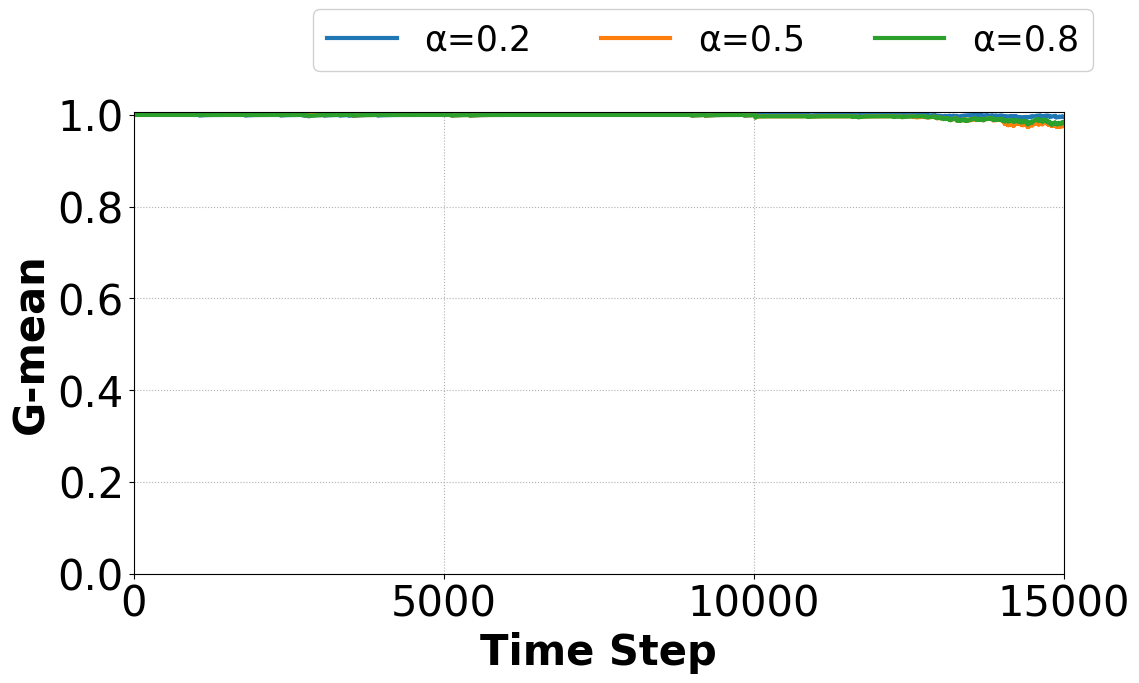}
      \caption{Blob}
      \label{fig:blob_alpha}
    \end{subfigure}%
    \begin{subfigure}{.33\textwidth}
      \centering
      \includegraphics[width=0.98\textwidth]{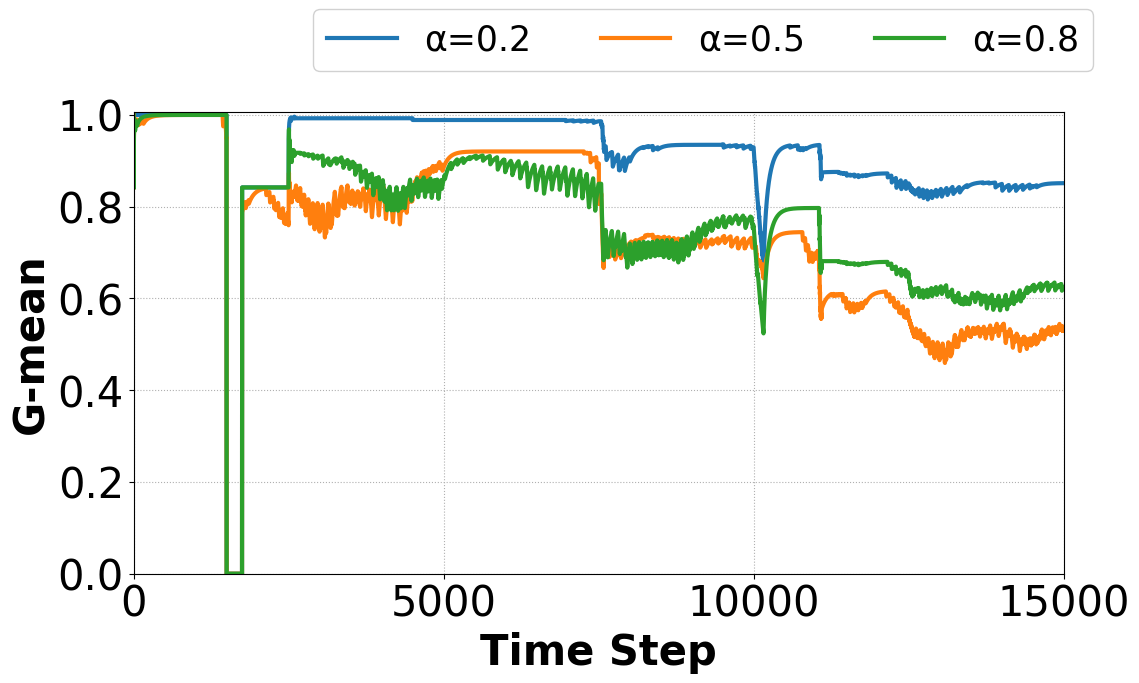} 
      \caption{Forest}
      \label{fig:forest_alpha}
    \end{subfigure}%
    \begin{subfigure}{.33\textwidth}
      \centering
      \includegraphics[width=0.98\textwidth]{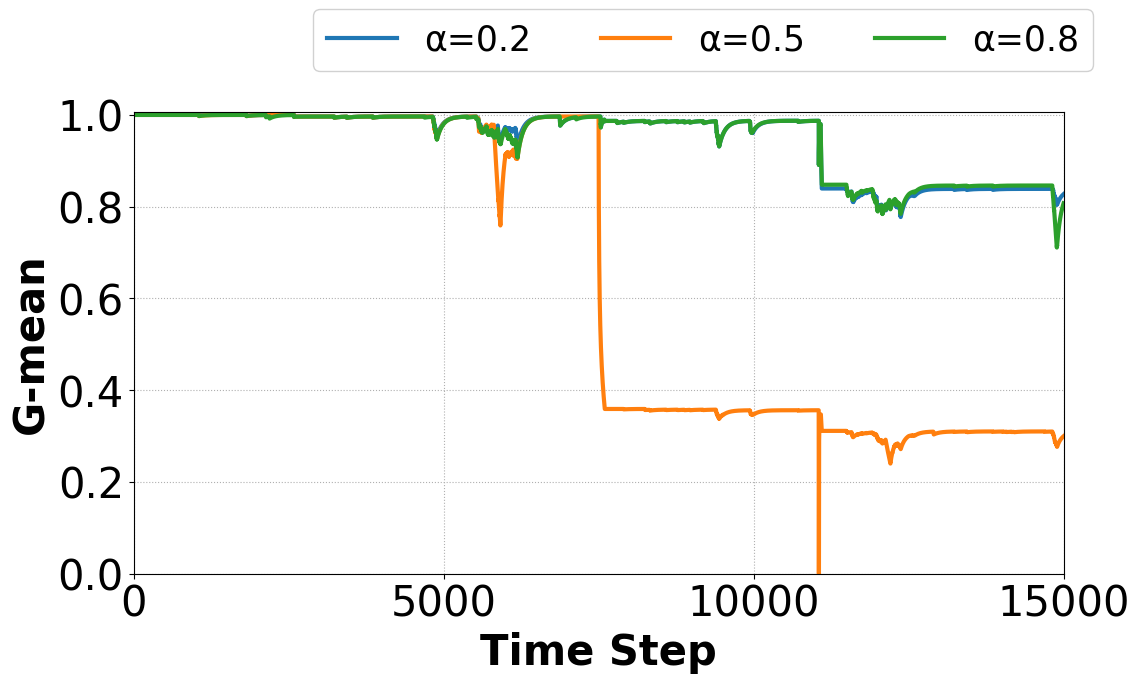} 
      \caption{KDD99}
      \label{fig:kdd_alpha}
    \end{subfigure}%
    
\caption{Comparison of different loss settings in nonstationary environments.}
\label{fig:loss_compare}
\end{figure*}

\begin{figure*}[!t]
 \begin{subfigure}{.33\textwidth} 
  \centering
  \includegraphics[width=0.98\textwidth]{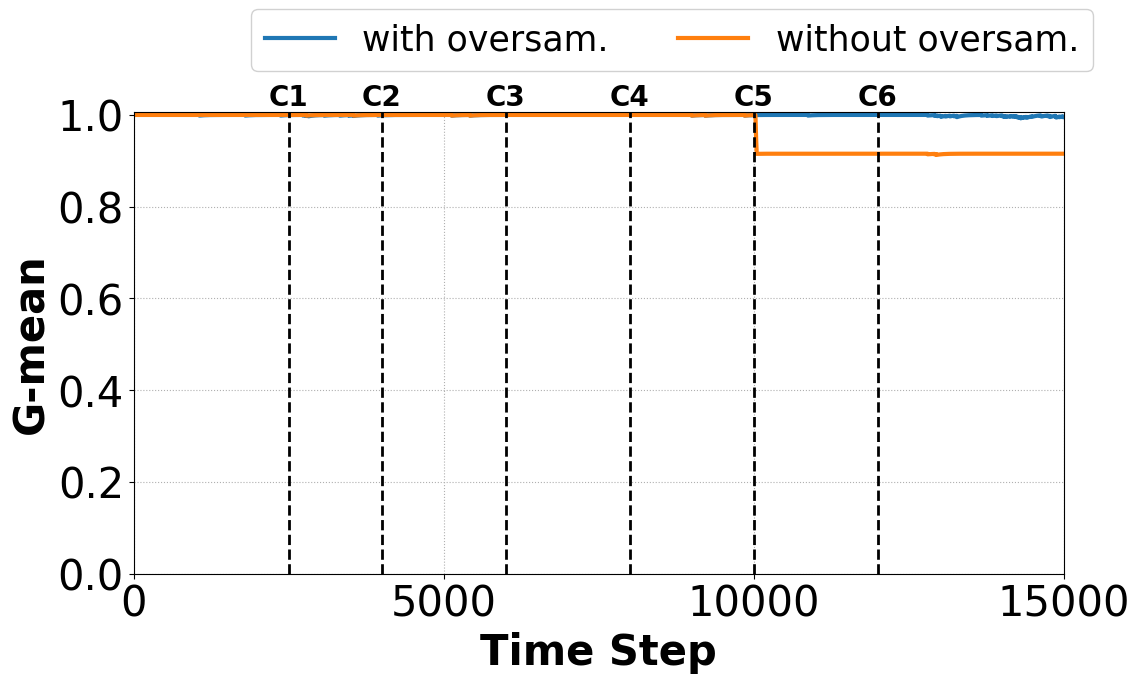} 
  \caption{Blob}
  \label{fig:blob_upsam}
 \end{subfigure}%
\begin{subfigure}{.33\textwidth}
  \centering
  \includegraphics[width=0.98\textwidth]{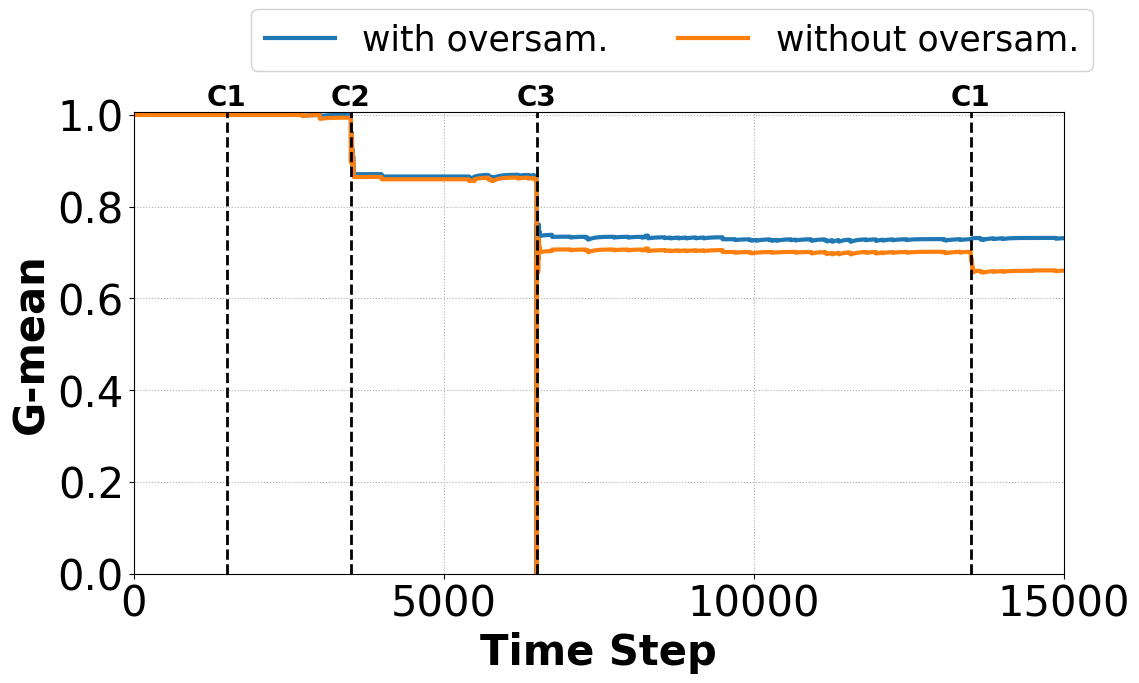}
  \caption{Sea}
  \label{fig:sea_upsam}
\end{subfigure}%
   \begin{subfigure}{.33\textwidth}
  \centering
  \includegraphics[width=0.98\textwidth]{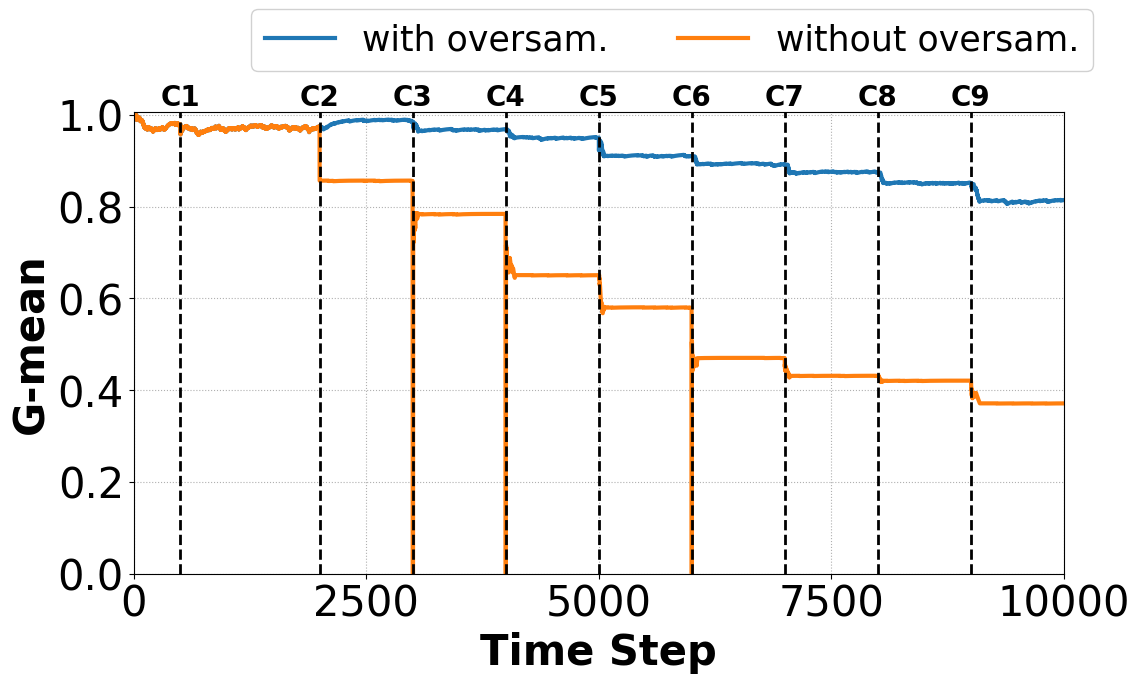} 
  \caption{MNIST}
  \label{fig:mnist_upsam}
 \end{subfigure}%
 
\caption{Comparison of model performance with and without oversampling.}
\label{fig:abla_upsam}
\end{figure*}

 

 

\begin{figure}[!t]
\begin{subfigure}{.49\columnwidth}
  \centering
\includegraphics[width=\columnwidth]{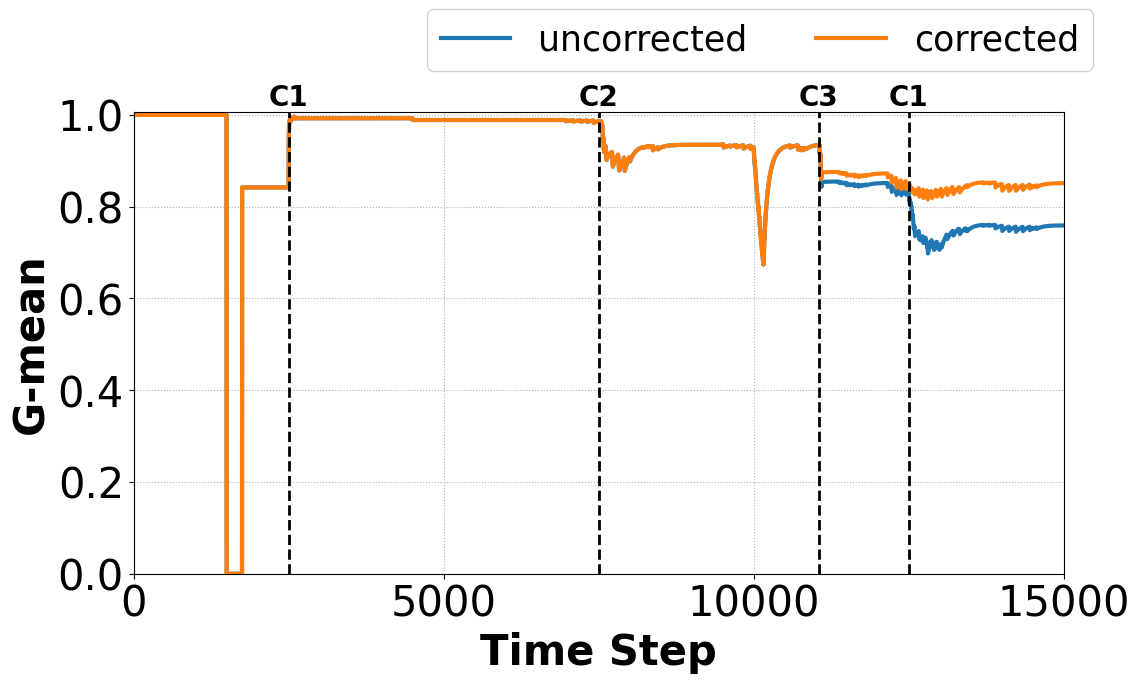}
  \caption{Forest}
\label{fig:forest_corr}
\end{subfigure}
\begin{subfigure}{.49\columnwidth}
  \centering
\includegraphics[width=\columnwidth]{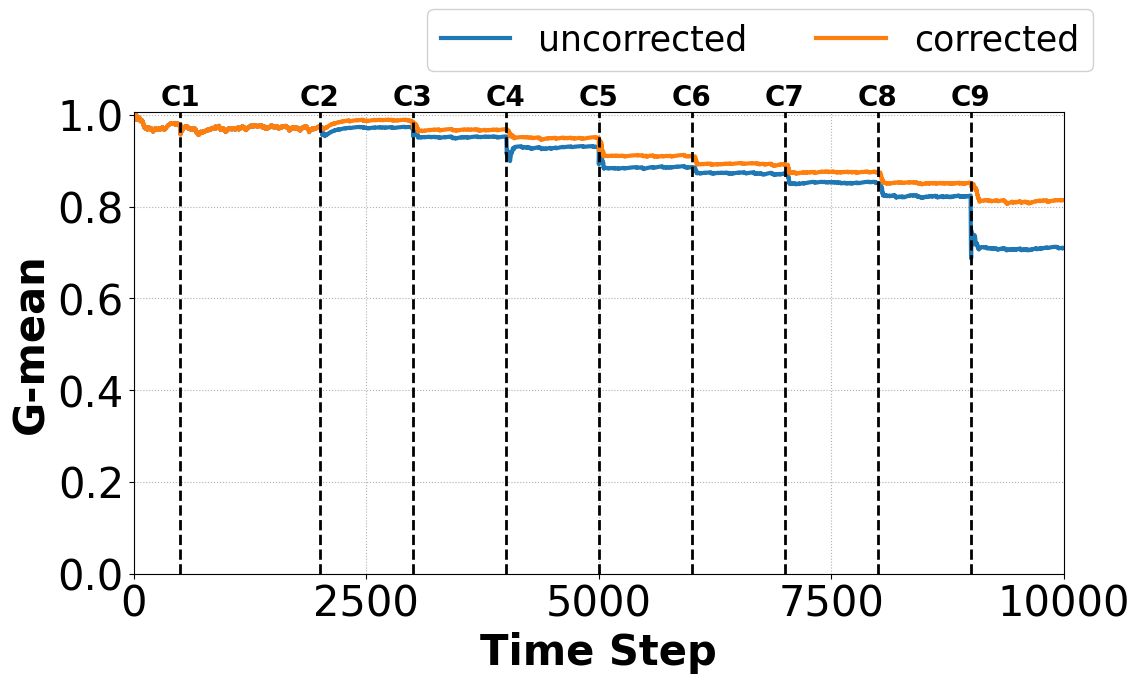}
  \caption{MNIST}
\label{fig:mnist_corr}
\end{subfigure}%
\caption{Comparison of G-mean with and without correction.}
\label{fig:correction}
\end{figure}

\begin{figure}[!t]
\begin{subfigure}{.49\columnwidth}
  \centering
\includegraphics[width=\columnwidth]{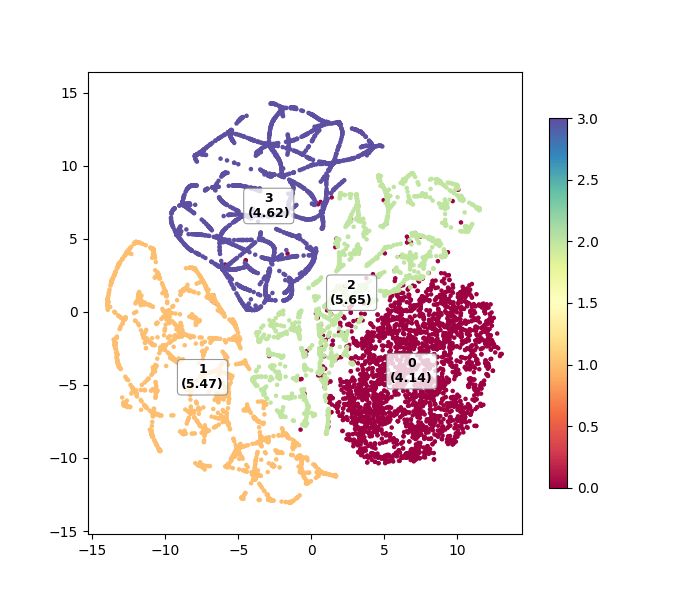}
  \caption{SMOTE after corrector}
\label{fig:mnist_tsne_corr}
\end{subfigure}
\begin{subfigure}{.49\columnwidth}
  \centering
\includegraphics[width=\columnwidth]{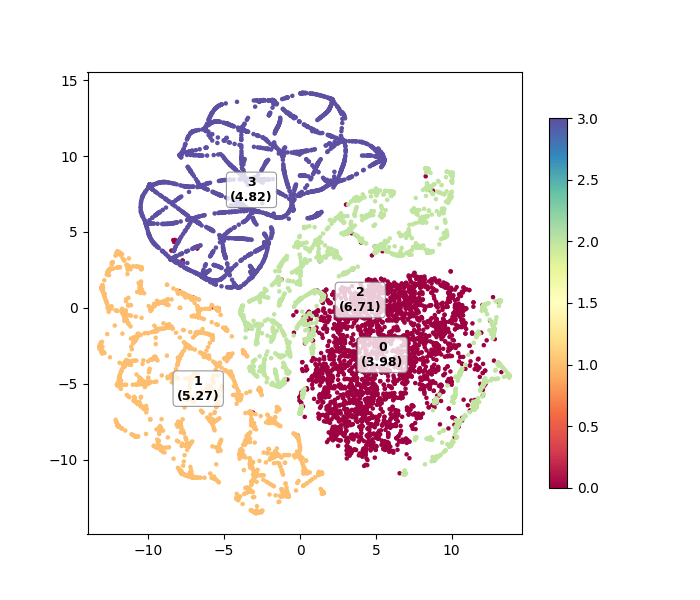}
  \caption{SMOTE without corrector}
\label{fig:mnist_tsne_uncorr}
\end{subfigure}%
\caption{Oversampled class distribution of MNIST at t=4000 with and without corrector, including the MDC value for each class.}
\label{fig:tsne_correction}
\end{figure}


\begin{figure}[!h]
	\centering \includegraphics[scale=0.32]{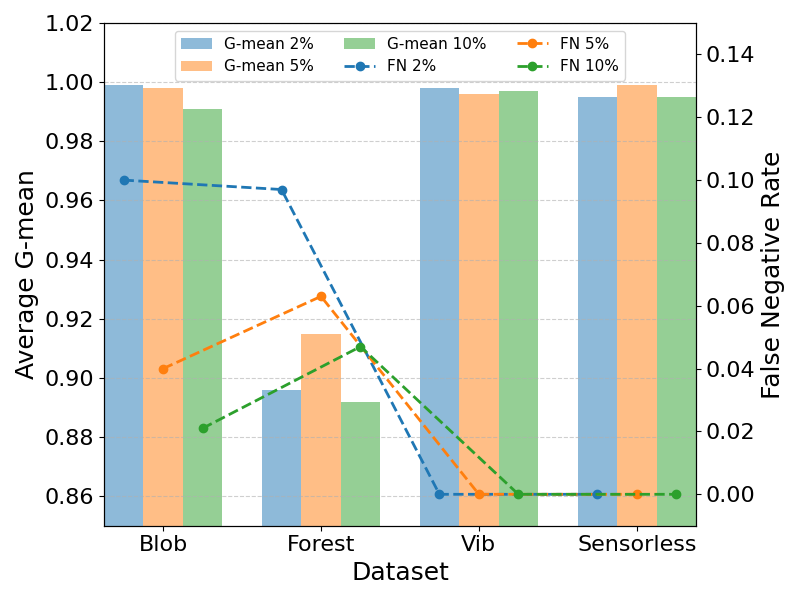}
	\caption{Comparison of different class imbalance rate.}
	\label{fig:imbalance}
\end{figure}

\subsubsection{Embeddings vs instance}

As shown in Fig.~\ref{fig:emb_compare}, embeddings significantly improve performance across all datasets. Using raw instances as input to the MLP causes a sharp performance drop when new classes appear, since the predictions and consequently the classification loss $L_{clf}$ depend entirely on the raw input. In contrast, embeddings influence the MLP output and are implicitly optimized through $L_{clf}$. When training with the combined loss $L_{total}$, both terms guide the learning of better representations. Replacing embeddings with raw instances removes this benefit, resulting in degraded performance, particularly on complex, high-dimensional datasets.

\subsubsection{Loss combination}
In this experiment, we conduct experiments with different combinations of $L_{recon}$ and $L_{clf}$ as $L_{total}$ to explore their influence on model performance. The combination of losses are shown in the equation below. Three combinations are considered: $\alpha=0.2$, $\alpha=0.5$, and $\alpha=0.8$. The equation of total loss can be found in Eq.~\ref{eq:total}. In this experiment, we use the Blob, Forest, and KDD99 datasets to demonstrate the results.

In Fig.~\ref{fig:loss_compare}, $\alpha=0.2$ yields slightly better performance on Blob and KDD99 and significantly better on Forest compared to $\alpha=0.8$, while $\alpha=0.5$ performs the worst across all datasets. This suggests that a balanced loss ($\alpha=0.5$) may lead to conflicting objectives, hindering feature learning. In contrast, an imbalanced loss prioritizes one objective, reducing interference between tasks and improving feature discrimination. $\alpha=0.2$ performs best, indicating that emphasizing classification helps learn more discriminative features while preserving reconstruction quality. Thus, we adopt $\alpha=0.2$ for the following experiments.

\subsubsection{Oversampling vs no oversampling}
In this experiment, we evaluate the necessity of oversampling for class imbalance. Without oversampling, the minority class remains unaltered during both new class and incremental training (Lines 19 and 21 in the pseudocode). As shown in Fig.~\ref{fig:abla_upsam}, models trained with oversampling consistently outperform those without, especially when introducing a new class. This underscores the importance of oversampling, as SMOTE reduces sparsity by generating interpolated samples, creating a more continuous and enriched feature distribution that helps the model capture key patterns effectively.

\subsubsection{Impact of the correction mechanism} In this experiment, we examine the necessity of the correction mechanism by comparing the model performance with and without the corrector. We provide the comparison illustrations of dataset Forest and MNIST in Fig.~\ref{fig:correction}. As observed in both datasets, incorporating the corrector reduces error propagation across incremental steps, resulting in superior classification performance. This advantage becomes increasingly evident as more classes are introduced, where error accumulation without correction would otherwise degrade performance. Moreover, the performance gap between the two settings varies across classes: in some categories, the corrector brings only minor improvements, while in others the gains are substantial. This discrepancy arises because classes that are highly similar to existing ones or have sparse/irregular distributions are more prone to misclassifications and error propagation, making the correction mechanism particularly effective for those cases.

In Fig.~\ref{fig:tsne_correction}, we illustrate how the corrector operates using the MNIST dataset at $t=4000$. We also compute the \textit{Mean Distance to Centroid} (MDC)\upcite{davies2009cluster} for each class in both figures, where a smaller MDC value indicates a higher intra-class compactness. As shown in Fig.~\ref{fig:mnist_tsne_uncorr}, without correction, class 2 forms a surrounding cluster around class 0 because misclassified class-0 instances are absorbed into class 2. This phenomenon is reflected by the increase in the MDC value of class 2. After applying the correction, while other classes remain slightly affected, class 2 exhibits the largest change in MDC, from 5.65 to 6.72, corresponding to the surrounding cluster observed in Fig.~\ref{fig:mnist_tsne_uncorr}. Oversampling such noisy instances amplifies the overlap and blurs the class boundary. In contrast, Fig.~\ref{fig:mnist_tsne_corr} shows that the corrector filters out outliers within each minority class based on density, repositioning class 2 adjacent to, rather than enclosing, class 0. This reduces boundary interference after SMOTE and enhances the overall class separability.

\subsubsection{Robustness to class imbalance}
In this experiment, we evaluate the average G-mean and false negative rate under class imbalance ratios of 2\%, 5\%, and 10\% using the Blob, Vib, Forest, and Sensorless datasets to assess the sensitivity of the proposed method to varying levels of imbalance. Under class imbalance, models tend to favor the majority class, leading to misclassification of minority class samples as majority and thereby increasing false negatives. For evaluation, we regard the majority class as the negative class and all minority classes as positive.

The false negative rates and average G-means are shown in Fig.~\ref{fig:imbalance}. Across all four datasets, the G-mean remains consistently high with only minor fluctuations, regardless of the imbalance ratio. Although increasing the number of minority samples can facilitate upsampling, it also risks amplifying noisy instances, which may distort the true class distribution. Furthermore, when new samples largely overlap with existing ones in the feature space, the benefits of upsampling diminish. Consequently, a higher quantity of minority samples does not always yield more effective upsampling, and its impact on performance may vary. Nevertheless, our model demonstrates consistently strong, stable, and resilient performance across both moderate and extreme imbalance scenarios, underscoring its robustness to class imbalance. Notably, false negative rates remain below 0.1 even in the most imbalanced case, and in some datasets drop to zero. Moreover, as the imbalance ratio increases, the false negative rate decreases, highlighting the potential of our method in more balanced settings.

\begin{figure}[!t]
  \begin{subfigure}{.5\columnwidth}
  \centering
\includegraphics[width=1.0\columnwidth]{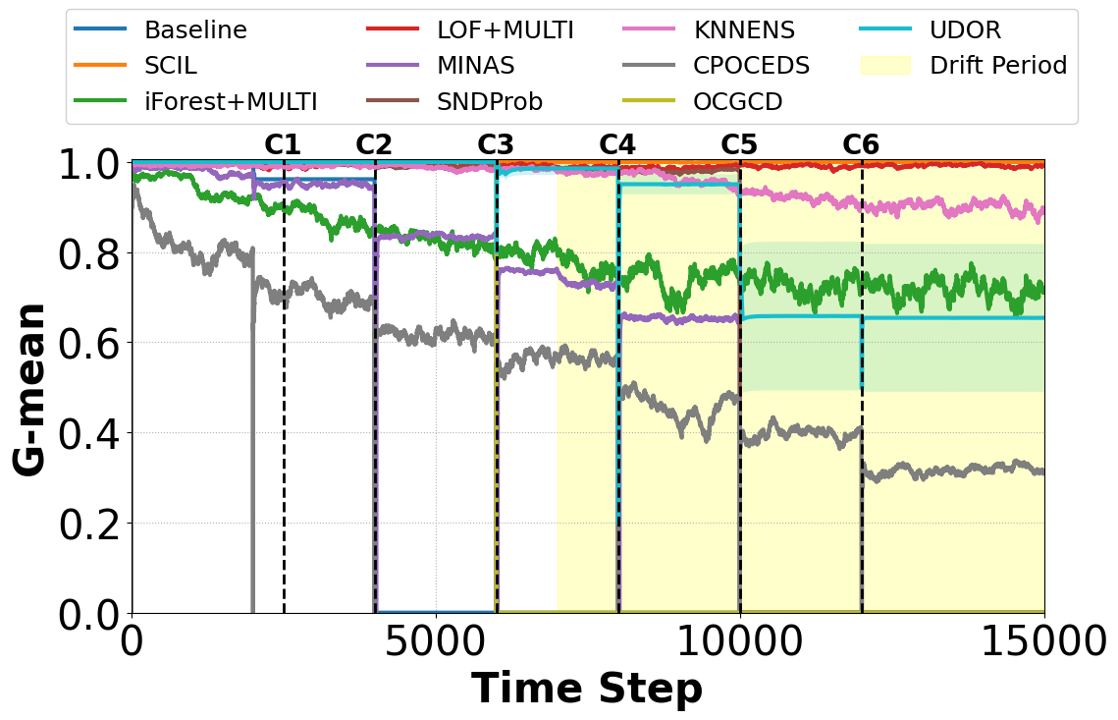} 
  \caption{Blob}
  \label{fig:blob_compare}
 \end{subfigure}%
  \begin{subfigure}{.5\columnwidth}
  \centering
\includegraphics[width=1.0\columnwidth]{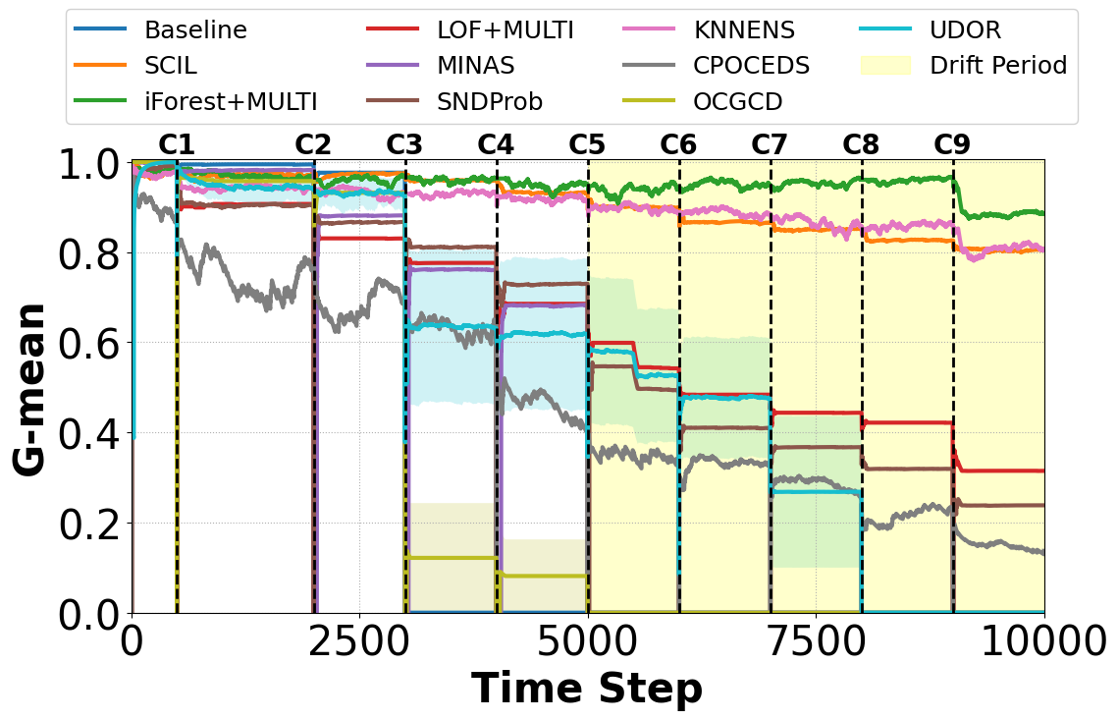} 
  \caption{MNIST}
  \label{fig:mnist_compare}
 \end{subfigure}%

  \begin{subfigure}{.5\columnwidth}
  \centering
\includegraphics[width=1.\columnwidth]{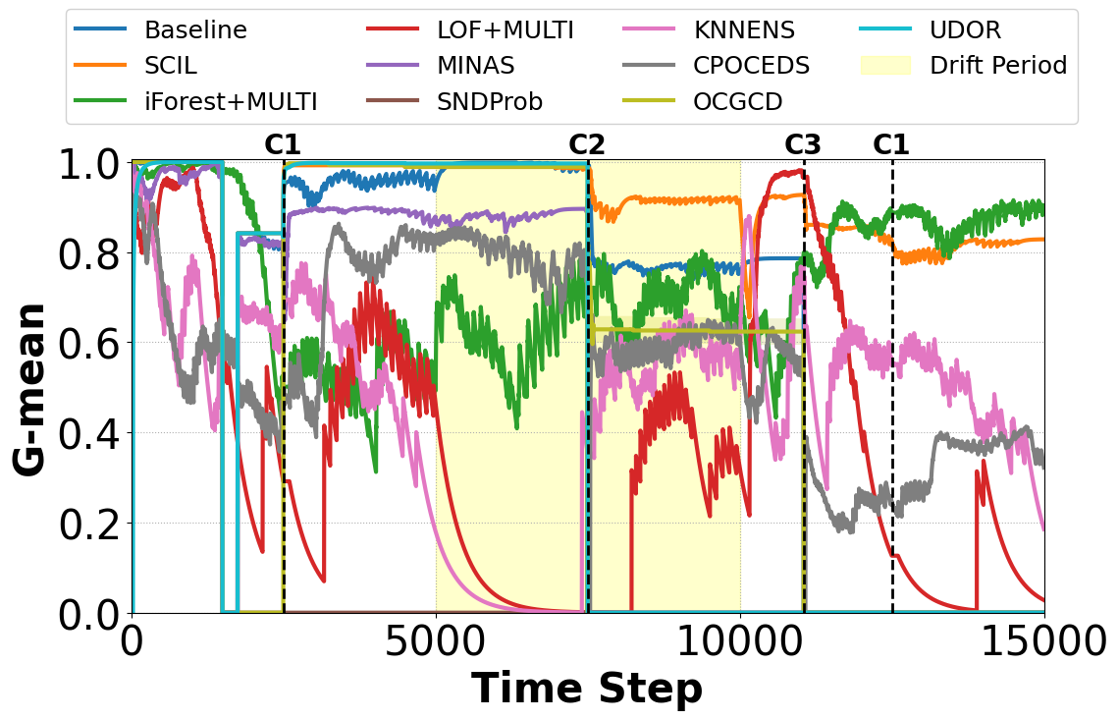} 
  \caption{Forest}
  \label{fig:forest_compare}
 \end{subfigure}%
   \begin{subfigure}{.5\columnwidth}
\centering\includegraphics[width=1.0\columnwidth]{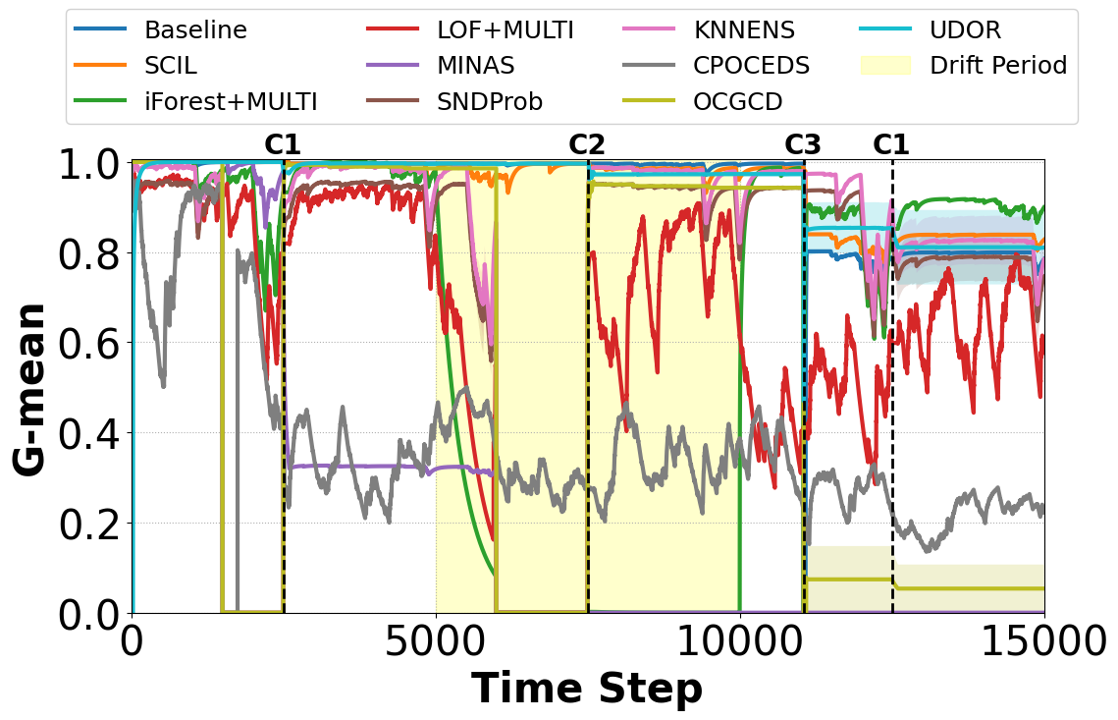} 
  \caption{KDD99}
  \label{fig:kdd_compare}
 \end{subfigure}%

\caption{G-mean performance of different methods.}
\label{fig:method_compare}
\end{figure}

\begin{table*}[!t]
\centering
\caption{Average EN\_Accuracy.\\(Best results are in \textbf{bold}, second-best are in \textcolor{blue}{\textbf{bold blue}}}
\label{tab:en_acc}

\begin{adjustbox}{width=0.99\textwidth}
\begin{tabular}{|c|ccc|cccccc|}
\hline
              & \multicolumn{3}{c|}{\textbf{Synthetic datasets}} 
              & \multicolumn{6}{c|}{\textbf{Real-world datasets}} \\ \hline
Methods       
& \multicolumn{1}{c|}{Sea} 
& \multicolumn{1}{c|}{Vib} 
& Blob 
& \multicolumn{1}{c|}{WDN} 
& \multicolumn{1}{c|}{MNIST} 
& \multicolumn{1}{c|}{KDD99} 
& \multicolumn{1}{c|}{Forest} 
& \multicolumn{1}{c|}{Sensorless} 
& Shuttle \\ \hline

Baseline      
& \multicolumn{1}{c|}{0.966(0.000)} 
& \multicolumn{1}{c|}{0.987(0.000)} 
& 0.987(0.000)
& \multicolumn{1}{c|}{0.889(0.000)} 
& \multicolumn{1}{c|}{\textcolor{blue}{\textbf{0.969(0.000)}}} 
& \multicolumn{1}{c|}{0.985(0.000)}
& \multicolumn{1}{c|}{\textcolor{blue}{\textbf{0.994(0.000)}}} 
& \multicolumn{1}{c|}{\textcolor{blue}{\textbf{0.993(0.000)}}} 
& \textcolor{blue}{\textbf{0.846(0.000)}} \\ \hline

iForest+MULTI 
& \multicolumn{1}{c|}{0.438(0.000)} 
& \multicolumn{1}{c|}{0.892(0.000)} 
& 0.449(0.000) 
& \multicolumn{1}{c|}{0.561(0.000)} 
& \multicolumn{1}{c|}{0.688(0.000)} 
& \multicolumn{1}{c|}{0.439(0.000)} 
& \multicolumn{1}{c|}{0.444(0.000)} 
& \multicolumn{1}{c|}{0.854(0.000)} 
& 0.591(0.000) \\ \hline

LOF+MULTI     
& \multicolumn{1}{c|}{0.928(0.000)} 
& \multicolumn{1}{c|}{0.978(0.000)} 
& 0.943(0.000) 
& \multicolumn{1}{c|}{0.708(0.000)} 
& \multicolumn{1}{c|}{0.949(0.000)} 
& \multicolumn{1}{c|}{0.895(0.000)} 
& \multicolumn{1}{c|}{0.903(0.000)} 
& \multicolumn{1}{c|}{0.933(0.000)} 
& 0.670(0.000) \\ \hline

MINAS         
& \multicolumn{1}{c|}{0.704(0.000)} 
& \multicolumn{1}{c|}{0.900(0.000)} 
& 0.800(0.000) 
& \multicolumn{1}{c|}{0.590(0.000)} 
& \multicolumn{1}{c|}{0.944(0.000)} 
& \multicolumn{1}{c|}{0.941(0.000)} 
& \multicolumn{1}{c|}{0.918(0.000)} 
& \multicolumn{1}{c|}{0.954(0.000)} 
& 0.676(0.000) \\ \hline

CPOCEDS       
& \multicolumn{1}{c|}{0.438(0.000)} 
& \multicolumn{1}{c|}{0.480(0.000)} 
& 0.536(0.000) 
& \multicolumn{1}{c|}{0.464(0.000)} 
& \multicolumn{1}{c|}{0.467(0.000)} 
& \multicolumn{1}{c|}{0.297(0.000)} 
& \multicolumn{1}{c|}{0.569(0.000)} 
& \multicolumn{1}{c|}{0.628(0.000)} 
& 0.604(0.000) \\ \hline

SNDProb       
& \multicolumn{1}{c|}{\textcolor{blue}{\textbf{0.993(0.000)}}} 
& \multicolumn{1}{c|}{\textcolor{blue}{\textbf{0.991(0.000)}}} 
& 0.943(0.039) 
& \multicolumn{1}{c|}{\textcolor{blue}{\textbf{0.987(0.000)}}} 
& \multicolumn{1}{c|}{0.506(0.000)} 
& \multicolumn{1}{c|}{0.964(0.000)} 
& \multicolumn{1}{c|}{0.981(0.000)} 
& \multicolumn{1}{c|}{0.961(0.000)} 
& 0.562(0.000) \\ \hline

KNNENS        
& \multicolumn{1}{c|}{\textbf{0.997(0.000)}} 
& \multicolumn{1}{c|}{0.745(0.055)} 
& 0.696(0.136) 
& \multicolumn{1}{c|}{0.758(0.009)} 
& \multicolumn{1}{c|}{0.705(0.000)} 
& \multicolumn{1}{c|}{0.848(0.000)} 
& \multicolumn{1}{c|}{0.303(0.052)} 
& \multicolumn{1}{c|}{0.961(0.003)} 
& 0.818(0.002) \\ \hline

OCGCD       
& \multicolumn{1}{c|}{\textbf{0.997(0.000)}} 
& \multicolumn{1}{c|}{0.990(0.001)} 
& \textbf{0.993(0.000)} 
& \multicolumn{1}{c|}{0.889(0.000)} 
& \multicolumn{1}{c|}{0.962(0.001)} 
& \multicolumn{1}{c|}{\textbf{0.996(0.000)}} 
& \multicolumn{1}{c|}{\textcolor{blue}{\textbf{0.994(0.000)}}} 
& \multicolumn{1}{c|}{0.991(0.001)} 
& 0.808(0.004) \\ \hline

UDOR       
& \multicolumn{1}{c|}{0.976(0.005)}
& \multicolumn{1}{c|}{0.990(0.002)} 
& \textcolor{blue}{\textbf{0.992(0.001)}}
& \multicolumn{1}{c|}{0.723(0.002)} 
& \multicolumn{1}{c|}{0.807(0.011)} 
& \multicolumn{1}{c|}{0.983(0.001)} 
& \multicolumn{1}{c|}{0.979(0.004)} 
& \multicolumn{1}{c|}{0.980(0.005)} 
& 0.840(0.014) \\ \hline

SCIL (Ours)   
& \multicolumn{1}{c|}{\textcolor{blue}{\textbf{0.993(0.000)}}} 
& \multicolumn{1}{c|}{\textbf{0.996(0.000)}} 
& \textbf{0.993(0.000)} 
& \multicolumn{1}{c|}{\textbf{0.993(0.000)}} 
& \multicolumn{1}{c|}{\textbf{0.970(0.000)}} 
& \multicolumn{1}{c|}{\textcolor{blue}{\textbf{0.989(0.000)}}} 
& \multicolumn{1}{c|}{\textbf{0.998(0.000)}} 
& \multicolumn{1}{c|}{\textbf{0.996(0.000)}} 
& \textbf{0.916(0.000)} \\ \hline
\end{tabular}
\end{adjustbox}
\end{table*}

\begin{table*}[!t]
\centering
\caption{Average G-mean.\\(Best results are in \textbf{bold}, second-best are in \textcolor{blue}{\textbf{bold blue}}}
\label{tab:gmean}

\begin{adjustbox}{width=0.99\textwidth}
\begin{tabular}{|c|ccc|cccccc|}
\hline
               & \multicolumn{3}{c|}{\textbf{Synthetic datasets}} 
               & \multicolumn{6}{c|}{\textbf{Real-world datasets}} \\ \hline
Methods        
& \multicolumn{1}{c|}{Sea} 
& \multicolumn{1}{c|}{Vib} 
& \multicolumn{1}{c|}{Blob} 
& \multicolumn{1}{c|}{WDN} 
& \multicolumn{1}{c|}{MNIST} 
& \multicolumn{1}{c|}{KDD99} 
& \multicolumn{1}{c|}{Forest} 
& \multicolumn{1}{c|}{Sensorless} 
& Shuttle \\ \hline

Baseline       
& \multicolumn{1}{c|}{0.423(0.000)} 
& \multicolumn{1}{c|}{0.833(0.000)} 
& \multicolumn{1}{c|}{0.273(0.000)} 
& \multicolumn{1}{c|}{\textcolor{blue}{\textbf{0.990(0.000)}}} 
& \multicolumn{1}{c|}{0.209(0.000)} 
& \multicolumn{1}{c|}{\textcolor{blue}{\textbf{0.949(0.000)}}}
& \multicolumn{1}{c|}{0.658(0.000)} 
& \multicolumn{1}{c|}{0.600(0.000)} 
& 0.466(0.000) \\ \hline

iForest+MULTI  
& \multicolumn{1}{c|}{0.660(0.000)} 
& \multicolumn{1}{c|}{0.976(0.000)} 
& \multicolumn{1}{c|}{0.805(0.000)} 
& \multicolumn{1}{c|}{0.534(0.000)} 
& \multicolumn{1}{c|}{\textbf{0.954(0.000)}} 
& \multicolumn{1}{c|}{0.646(0.000)} 
& \multicolumn{1}{c|}{\textcolor{blue}{\textbf{0.736(0.000)}}} 
& \multicolumn{1}{c|}{0.502(0.000)} 
& \textcolor{blue}{\textbf{0.704(0.000)}} \\ \hline

LOF+MULTI      
& \multicolumn{1}{c|}{\textbf{0.956(0.000)}} 
& \multicolumn{1}{c|}{\textbf{0.994(0.000)}} 
& \multicolumn{1}{c|}{\textcolor{blue}{\textbf{0.991(0.000)}}} 
& \multicolumn{1}{c|}{0.703(0.000)} 
& \multicolumn{1}{c|}{0.642(0.000)} 
& \multicolumn{1}{c|}{0.642(0.000)} 
& \multicolumn{1}{c|}{0.376(0.000)} 
& \multicolumn{1}{c|}{0.588(0.000)} 
& 0.688(0.000) \\ \hline

MINAS          
& \multicolumn{1}{c|}{0.568(0.000)} 
& \multicolumn{1}{c|}{0.889(0.000)} 
& \multicolumn{1}{c|}{0.794(0.000)} 
& \multicolumn{1}{c|}{0.433(0.000)} 
& \multicolumn{1}{c|}{0.760(0.000)} 
& \multicolumn{1}{c|}{0.249(0.000)} 
& \multicolumn{1}{c|}{0.436(0.000)} 
& \multicolumn{1}{c|}{0.585(0.000)} 
& 0.548(0.000) \\ \hline

CPOCEDS        
& \multicolumn{1}{c|}{0.321(0.000)} 
& \multicolumn{1}{c|}{0.232(0.000)} 
& \multicolumn{1}{c|}{0.221(0.000)} 
& \multicolumn{1}{c|}{0.473(0.000)} 
& \multicolumn{1}{c|}{0.164(0.000)} 
& \multicolumn{1}{c|}{0.368(0.000)} 
& \multicolumn{1}{c|}{0.389(0.000)} 
& \multicolumn{1}{c|}{0.517(0.000)} 
& 0.579(0.000) \\ \hline

SNDProb        
& \multicolumn{1}{c|}{\textcolor{blue}{\textbf{0.873(0.000)}}} 
& \multicolumn{1}{c|}{0.872(0.000)} 
& \multicolumn{1}{c|}{0.669(0.001)} 
& \multicolumn{1}{c|}{0.983(0.000)} 
& \multicolumn{1}{c|}{0.613(0.000)} 
& \multicolumn{1}{c|}{0.741(0.000)} 
& \multicolumn{1}{c|}{0.101(0.000)} 
& \multicolumn{1}{c|}{0.200(0.000)} 
& 0.578(0.000) \\ \hline

KNNENS         
& \multicolumn{1}{c|}{0.430(0.000)} 
& \multicolumn{1}{c|}{0.919(0.010)} 
& \multicolumn{1}{c|}{0.957(0.004)} 
& \multicolumn{1}{c|}{0.986(0.002)} 
& \multicolumn{1}{c|}{0.915(0.004)} 
& \multicolumn{1}{c|}{0.772(0.008)} 
& \multicolumn{1}{c|}{0.534(0.030)} 
& \multicolumn{1}{c|}{0.938(0.001)} 
& 0.670(0.001) \\ \hline

OCGCD       
& \multicolumn{1}{c|}{0.870(0.030)} 
& \multicolumn{1}{c|}{0.970(0.012)} 
& 0.876(0.006) 
& \multicolumn{1}{c|}{0.763(0.002)} 
& \multicolumn{1}{c|}{0.538(0.007)} 
& \multicolumn{1}{c|}{0.943(0.014)}
& \multicolumn{1}{c|}{0.471(0.001)}
& \multicolumn{1}{c|}{0.200(0.001)} 
& 0.490(0.003) \\ \hline

UDOR       
& \multicolumn{1}{c|}{0.742(0.020)} 
& \multicolumn{1}{c|}{0.860(0.030)} 
& 0.400(0.000) 
& \multicolumn{1}{c|}{0.216(0.000)} 
& \multicolumn{1}{c|}{0.307(0.015)} 
& \multicolumn{1}{c|}{0.570(0.016)}
& \multicolumn{1}{c|}{0.579(0.007)}
& \multicolumn{1}{c|}{\textcolor{blue}{\textbf{0.940(0.005)}}} 
& 0.413(0.010) \\ \hline

SCIL (Ours)    
& \multicolumn{1}{c|}{0.871(0.000)} 
& \multicolumn{1}{c|}{\textcolor{blue}{\textbf{0.993(0.000)}}} 
& \multicolumn{1}{c|}{\textbf{0.992(0.000)} }
& \multicolumn{1}{c|}{\textbf{0.992(0.000)}} 
& \multicolumn{1}{c|}{\textcolor{blue}{\textbf{0.920(0.000)}}} 
& \multicolumn{1}{c|}{\textbf{0.954(0.000)}} 
& \multicolumn{1}{c|}{\textbf{0.914(0.000)}} 
& \multicolumn{1}{c|}{\textbf{0.999(0.000)}} 
& \textbf{0.824(0.000)} \\ \hline
\end{tabular}
\end{adjustbox}
\end{table*}

\subsection{Comparative study}

This section compares the performance of seven methods—Baseline, MINAS, CPOCEDS, SCIL, iForest+MULTI, LOF+MULTI, SNDProb, KNNENS, OCGCD, and UDOR—using EN\_Accuracy and G-mean. Tables~\ref{tab:en_acc} and~\ref{tab:gmean} report average EN\_Accuracy and G-mean, highlighting the best methods in bold and the second-best are in bold blue. Fig.~\ref{fig:method_compare} shows G-mean over time.

As shown in both tables, the proposed SCIL method consistently ranks among the top three across all datasets in Table~\ref{tab:gmean} and among the top two in Table~\ref{tab:en_acc}, demonstrating its accuracy and robustness. In contrast, other methods show notable variability. For instance, LOF+MULTI performs well on Sea, Blob, and Vib but poorly on others.

EN\_Accuracy and G-mean do not always align. For instance, SNDProb shows low G-mean but high EN\_Accuracy on KDD99 and Forest due to good performance on seen classes, which dominate due to class imbalance. In such cases, $A_o$ heavily influences EN\_Accuracy. A low EN\_Accuracy implies poor overall classification, even without new classes, while high G-mean with low EN\_Accuracy suggests strong new class detection but weak seen-class classification—seen with iForest+MULTI, LOF+MULTI, and KNNENS. By contrast, Baseline and SCIL maintain high values for both metrics, evidencing balanced performance on seen and unseen classes.

As seen in Fig.~\ref{fig:method_compare}, except for KDD99, the G-mean of the Baseline drops significantly when facing new faults, indicating its inability to detect novel events. In contrast, SCIL, leveraging incremental learning, effectively addresses this issue by adapting to evolving data distributions, with strong performance on Blob and MNIST highlighting its strength in new class detection and incremental class management.

Further supporting these findings, Fig.~\ref{fig:forest_compare} and Fig.~\ref{fig:kdd_compare} show that concept drift causes severe performance collapse in existing methods: KNNENS and LOF+MULTI drop to zero after the first drift, as do iForest+MULTI and LOF+MULTI. These models misclassify drifted majority-class samples as minority classes, highlighting their inflexibility in non-stationary environments. SCIL, however, consistently preserves high G-mean through continuous adaptation, proving its resilience against concept drift.

Additionally, as observed in Fig.~\ref{fig:method_compare}, SCIL experiences a noticeable drop after encountering unseen classes (except on dataset Blob), which is also evident in other methods, such as in Fig.~\ref{fig:mnist_compare}. This phenomenon is due to the calculation of G-mean, which includes more recall values \( R_{i} \) as new classes are introduced. Since each \( R_{i} \) is less than or equal to 1, the product tends to decrease as \( n \) (the number of classes) increases. G-mean, being the geometric mean of recall values across all classes as shown in Eq.~\ref{eq:gmean}, may therefore decrease slightly with each added class, even if the recall for the new class is close to 1. Additionally, G-mean depends on classification performance over time; thus, if the model’s classification ability on known classes is low, the G-mean remains low until new classes arrive.

In summary, SCIL demonstrates its strongest advantages on high-dimensional datasets and in non-stationary environments with multiple concept drifts, where existing methods often suffer from severe performance degradation. Specifically, the experimental results show that SCIL achieves effective performance recovery under recurrent drift scenarios and maintains competitive performance on high-dimensional image datasets.

\section{Conclusion}\label{sec:conclusion}
Classification with data streams faces major challenges, including nonstationary environments, label unavailability, class imbalance, and incremental new classes. To address these issues, we propose SCIL, a unified and resilient framework that integrates an AE with an MLP. The MLP performs multi-class classification, while the AE’s reconstruction loss supports new class detection, and a classification corrector mitigates error propagation. SCIL further employs a queue mechanism to store samples of both seen and unseen classes for future updates, leveraging incremental learning to adapt to concept drift. To evaluate robustness under class imbalance, we simulate realistic scenarios comprising one majority class and multiple minority classes, combined with various concept drift patterns. Furthermore, we investigate the model's resilience and stability across different class imbalance ratios in the ablation study. We also conduct extensive ablation studies on input configurations, loss functions, drift detection mechanisms, oversampling strategies, and the correction module to justify the design choices. Experimental results demonstrate that SCIL consistently and significantly outperforms both baseline and advanced methods across varied datasets.

	\bibliographystyle{gb7714-2015} 
	\bibliography{refs}

\end{document}